%% file: main.tex
\documentclass{article}

  \PassOptionsToPackage{numbers, compress}{natbib}

\usepackage{amsmath,amssymb,graphicx,xcolor,algorithm,algpseudocode,booktabs,bm,relsize,enumitem,multirow,amsthm,subfigure,epsfig, caption}
\definecolor{Blue9}{rgb}{0.098,0.3,0.9}
\usepackage[colorlinks=true,linkcolor=Blue9,citecolor=Blue9]{hyperref}

\usepackage[preprint]{neurips_2020}

\newcommand{\reviewer}[3]{
	\expandafter\newcommand\csname #1\endcsname[1]{
		\textcolor{#3}{#2: ##1}
	}
}
\reviewer{kimin}{KM}{red}
\reviewer{misha}{misha}{blue}
\reviewer{aravind}{AS}{cyan}
\reviewer{lhao}{HL}{orange}

\def\mytitle{Balancing Value Underestimation and Overestimation with Realistic Actor-Critic}

\usepackage[utf8]{inputenc} 
\usepackage[T1]{fontenc}  
\usepackage{hyperref}    
\usepackage{url}      
\usepackage{booktabs}    
\usepackage{amsfonts}    
\usepackage{nicefrac}    
\usepackage{microtype}   

\title{\mytitle}

\author{%
  Sicen Li, Gang Wang, Qinyun Tang, Liquan Wang \\ 
  Harbin Engineering University\\
 \texttt{\{ihuhuhu, wanggang, 993740540, wangliquan\}@hrbeu.edu.cn} \\
}

\begin{document}

\maketitle

\begin{abstract}
Model-free deep reinforcement learning (RL) has been successfully applied to challenging continuous control domains. 
However, poor sample efficiency prevents these methods from being widely used in real-world domains.
This paper introduces a novel model-free algorithm, Realistic Actor-Critic(RAC), which can be incorporated with any off-policy RL algorithms to improve sample efficiency.
RAC employs Universal Value Function Approximators (UVFA) to simultaneously learn a policy family with the same neural network, each with different trade-offs between underestimation and overestimation.
To learn such policies, we introduce uncertainty punished Q-learning, which uses uncertainty from the ensembling of multiple critics to build various confidence-bounds of Q-function.
We evaluate RAC on the MuJoCo benchmark, achieving 10x sample efficiency and 25\% performance improvement on the most challenging Humanoid environment compared to SAC. 
\end{abstract}

\section{Introduction}
Sample efficiency is one of the main challenges that prevent reinforcement learning(RL) applying to real-world systems~\cite{dulac2020empirical, sutton2018reinforcement}. Recently, in continuous control domains, model-free off-policy reinforcement learning(RL) method has achieved a comparable sample efficiency to model-based methods by training accurate value approximations with a high Update-To-Data (UTD) ratio~\cite{chen2021randomized}. Quality of value approximation is a key for sample efficiency, stability and final performance as policy optimization relies on gradients of Q-functions to provide the direction of the policy update.

Overestimation bias and accumulation of function approximation errors in temporal difference methods~\cite{thrun1993issues, pendrith1997estimator, fujimoto2018addressing} are some of the main factors that plague value approximation. Undesirable overestimation bias may lead to sub-optimal policy updates and divergent behavior.


One way to address above issues is using ensemble methods~\cite{lan2020maxmin, fujimoto2018addressing, lee2020sunrise, chen2021randomized, kuznetsov2020controlling} to introduce underestimation bias which does not tend to be propagated during learning, as actions with low value estimates are avoided by the policy. However, underestimation bias may harm exploration by causing a pessimistic underexploration problem~\cite{ciosek2019better}. Both under- and overestimation bias may improve learning performance, depending on the environment~\cite{lan2020maxmin}. To overcome this problem, carefully adjusted hyperparameters are needed to trade off between under- and overestimation.

In this paper, we propose Realistic Actor-Critic (RAC) to address this under-overestimation trade-off, whose main idea is to learn together diverse policies with respect to various confidence-bounds of Q-functions in the same network.
In such a way, policies guided by upper confidence bounds (UCB) generate effective exploratory behaviors without falling in pessimistic underexploration, while other policies benefit from lower-confidence bounds(LCB) to control overestimation bias to provide consistency and stable convergence. We propose to jointly learn such family of policies parameterized with the Universal Value Function Approximators (UVFA)~\cite{schaul2015universal}. The best policy can be found by evaluating the learned policies at the evaluation phase. The learning process can be considered as a set of auxiliary tasks~\cite{badia2020never, lyle2021effect} that help build shared state representations and sills. 

However, learning such policies with diverse behaviors in a single network is challenging. We introduce uncertainty punished Q-learning(UPQ), which calculates uncertainty as punishment to correct value estimations. UPQ provides fine-granular overestimation control to make value approximation smoothly shifts from upper bounds to lower bounds. With UPQ, RAC incorporates various bounds of Q-function into a critic with UVFA to update policies that change smoothly. We propose to learn an ensemble of multiple critics that produces well-calibrated uncertainty estimations (i.e., standard deviation) on unseen samples~\cite{lee2020sunrise, pathak2019self, amos2018learning}. We show empirically that RAC controls the std and the mean of value estimate bias to close to zero for most of the training. Benefit from well value estimation, critics are trained with a high UTD ratio to improve sample efficiency significantly.

Empirically, we evaluate RAC combined with SAC~\cite{haarnoja2018soft} and TD3~\cite{fujimoto2018addressing} in continuous control benchmarks (OpenAI Gym~\cite{brockman2016openai}, MuJoCo~\cite{todorov2012mujoco}). Numerical results suggest RAC outperforms the current state of the art algorithms (MBPO~\cite{janner2019trust}, REDQ~\cite{chen2021randomized} and TQC~\cite{kuznetsov2020controlling}). RAC demonstrates 10x sample efficiency and 25\% performance improvement on the most challenging Humanoid environment compared to SAC. We perform ablations and isolate the effect of the main components of RAC on performance. Moreover, we perform hyperparameter ablations and demonstrate that RAC is stable in practice.

\section{Related work} \label{sec:relatedwork}

{\bf Underestimation and overestimation of Q-function}.
The maximization update rule in Q-learning has been shown to suffer from overestimation bias which will seriously hinder learning~\cite{thrun1993issues}. 

Minimization of a value ensemble is a common method to deal with  overestimation bias. Clipped double Q-learning (CDQ)~\cite{fujimoto2018addressing} takes the minimum value between a pair of critics to limit overestimation. SAC~\cite{haarnoja2018soft} then combined CDQ with entropy maximization to get impressive performance in continuous control tasks. Maxmin Q-learning~\cite{lan2020maxmin} mitigated the overestimation bias by using a minimization over multiple action-value estimates. But minimize a Q-function set is unable to filter out abnormally small values which causes undesired pessimistic underexploration problem~\cite{ciosek2019better}. Using minimization to control overestimation is coarse and wasteful as it ignores all estimates except the minimal one~\cite{kuznetsov2020controlling}. 

REDQ~\cite{chen2021randomized} proposed in-target minimization which used a minimization across a random subset of Q functions from the ensemble to alleviate the above problems. REDQ~\cite{chen2021randomized} showed their method reduces the std of the Q-function bias to close to zero for most of the training. Truncated Quantile Critics (TQC)~\cite{kuznetsov2020controlling} truncates the right tail of the distributional value ensemble by dropping several of the topmost atoms to control overestimation. Weighted bellman backups~\cite{lee2020sunrise} and  uncertainty weighted actor-critic~\cite{wu2021uncertainty} prevents error propagation~\cite{kumar2020discor} in Q-learning by reweighing sample transitions based on uncertainty estimations from the ensembles~\cite{lee2020sunrise} or Monte Carlo dropout~\cite{wu2021uncertainty, srivastava2014dropout}. 
Different from prior works, our work does not reweight sample transitions but directly adds uncertainty estimations to punish the target value. 

~\citet{lan2020maxmin} showed the effect of underestimation bias on learning efficiency is environment-dependent. It may be hard to choose the right parameters to balance under- and overestimation for completely different environments. Our work proposed to solve this problem by learning an optimistic and pessimistic policy family.

{\bf Ensemble methods}. 
In deep learning, ensemble methods often used to solve the two key issues, uncertainty estimations~\cite{wen2020batchensemble, abdar2021review} and out-of-distribution robustness~\cite{havasi2020training, dusenberry2020efficient, wenzel2020hyperparameter}. In reinforcement learning, using ensemble to enhance value function estimation was widely studied, such as, averaging a Q-ensemble~\cite{anschel2017averaged, peer2021ensemble}, bootstrapped actor-critic architecture~\cite{kalweit2017uncertainty, zheng122018self}, calculate uncertainty to reweight sample transitions~\cite{lee2020sunrise}, minimization over ensemble estimates~\cite{lan2020maxmin, chen2021randomized} and update the actor with a value ensemble~\cite{kuznetsov2020controlling, chen2021randomized}. 

A high-level policy can be distilled from a policy ensemble ~\cite{chen2019off, badia2020agent57} by density-based selection~\cite{saphal2020seerl}, selection through elimination~\cite{saphal2020seerl}, choosing action that max all Q-functions ~\cite{lee2020sunrise, parker2020effective,jung2020population}, Thompson Sampling~\cite{parker2020effective} and sliding-window UCBs~\cite{badia2020agent57}. Leveraging uncertainty estimations of the ensemble, ~\cite{osband2016deep, kalweit2017uncertainty, zheng122018self} simulated training different policies with a multi-head architecture independently to generate diverse exploratory behaviors. 

Ensemble methods were also used to learn joint state presentation to improve sample efficiency. There were two main methods: multi-heads ~\cite{osband2016deep, kalweit2017uncertainty, zheng122018self, goyal2019reinforcement} and UVFA~\cite{schaul2015universal, badia2020never, badia2020agent57}. In this paper, we uses uncertainty estimation to reduce value overestimation bias, a simple max strategy to to get the best policy and learning joint state presentation with UVFA. 

{\bf Optimistic exploration}. 
Pessimistic initialisation~\cite{rashid2020optimistic} and learning policy that maximizes a lower-confidence bound value could suffer pessimistic underexploration problem~\cite{ciosek2019better}. Optimistic exploration is a promising solution to ease the above problem by applying the principle of optimism in the face of uncertainty~\cite{brafman2002r}. Disagreement~\cite{pathak2019self} and EMI~\cite{kim2018emi} considered uncertainty as intrinsic motivation to encourage agent to explore the high uncertainty areas of the environment. Uncertainty punishment proposed in this paper can also be thought of as a special intrinsic motivation. Different with~\cite{pathak2019self, kim2018emi} which usually choose the weighting $\ge 0$ to encourage exploration, UPQ using the weighting $\le 0$ to control value bias. 

SUNRISE~\cite{lee2020sunrise} proposed an optimistic exploration that chooses the action that maximizes an upper-confidence bound (UCB)~\cite{chen2017ucb} of Q-functions. OAC~\cite{ciosek2019better} proposed an off-policy exploration strategy that is adjusted to a linear fit of UCB to the critic with the maximum KL divergence constraining between the exploration policy and the target policy. 

Most importantly, our work provides a unified framework for the under-overestimation trade-off.

\section{RAC}
We present Realistic Actor-Critic (RAC) which can be used in conjunction with most modern off-policy actor-critic RL algorithms in principle, such as SAC~\cite{haarnoja2018soft} and TD3~\cite{fujimoto2018addressing}. For the exposition, we describe only the SAC version of RAC (RAC-SAC) in the main body. The TD3 version of RAC (RAC-TD3) follows the same principles and is fully described in Appendix~\ref{appendix:setup}. 

\subsection{Problem setting and preliminaries}


We consider the standard reinforcement learning notation, with states $\mathbf{s}$, actions $\mathbf{a}$, reward $r(\mathbf{s},\mathbf{a})$, and dynamics $p(\mathbf{s'}\mid \mathbf{s},\mathbf{a})$. The discounted return $R_t =  {\textstyle \sum_{k=0}^{\infty }} \gamma^{k} r_{k}$ is the total accumulated rewards from timestep $t$, $\gamma \in [0, 1]$ is a discount factor determining the priority of short-term rewards. The objective is to find the optimal policy $\pi _{\phi}(\mathbf{s}\mid \mathbf{a})$ with parameters $\phi$, which maximizes the expected return $J(\phi)=\mathbb{E}_{p_\pi}\left[R_{t}\right]$.

The maximum entropy objective~\cite{ziebart2010modeling} encourages the robustness to noise and exploration by maximizing a weighted objective of the reward and the policy entropy:
\begin{align}
\pi^{*}=\arg \max _{\pi} \sum_{t} \mathbb{E}_{\mathbf{s} \sim p, \mathbf{a} \sim \pi}\left[r(\mathbf{s},\mathbf{a})+\alpha \mathcal{H}\left(\pi\left(\cdot \mid \mathbf{s}\right)\right)\right], \label{eq:maximumentropyobjective}
\end{align}
where $\alpha$ is the temperature parameter that can be used to determine the relative importance of entropy and reward. Soft Actor-Critic(SAC)~\cite{haarnoja2018soft} seeks to optimize the maximum entropy objective by alternating between a soft policy evaluation and a soft policy improvement. A parameterized soft Q-function $Q_{\theta}\left(\mathbf{s},\mathbf{a}\right)$ , known as the critic in actor-critic methods, is trained by minimizing the soft Bellman residual:
\begin{align}
  & \mathcal{L}_{\tt critic} (\theta) = \mathbb{E}_{\tau \sim \mathcal{B}} [\left(Q_\theta(\mathbf{s},\mathbf{a}) - y \right)^2], \label{eq:sac_critic_tot}
  \\
  & y = r - \gamma \mathbb{E}_{\mathbf{a'}\sim \pi_\phi} \big[ Q_{\bar \theta} (\mathbf{s'},\mathbf{a'}) - \alpha \log \pi_{\phi} (\mathbf{a'} \mid \mathbf{s'}) \big], \label{eq:sac_critic_target}
\end{align}
where $\tau = (\mathbf{s},\mathbf{a},r,\mathbf{s'})$ is a transition, 
$\mathcal{B}$ is a replay buffer, 
$\bar \theta$ are the delayed parameters which is updated by exponential moving average $\bar \theta \leftarrow \rho \theta+(1-\rho) \bar \theta$, $\rho$ is the target smoothing coefficient, $y$ is the target value.

The parameterized policy $\pi_\phi$ , known as the actor, is updated by minimizing the following object:
\begin{align}
  & \mathcal{L}_{\tt actor} (\phi) = \mathbb{E}_{\mathbf{s} \sim \mathcal{B}, \mathbf{a} \sim \pi_{\phi}}\left[\alpha \log \left(\pi_{\phi}\left(\mathbf{a}\mid\mathbf{s}\right)\right)-Q_{\theta}\left(\mathbf{a}, \mathbf{s}\right)\right]. \label{eq:sac_actor_tot}
\end{align}
SAC uses a automate entropy adjusting mechanism to update $\alpha$ with following objective:
\begin{align}
  & \mathcal{L}_{\tt temp} (\alpha) = \mathbb{E}_{\mathbf{s} \sim \mathcal{B}, \mathbf{a} \sim \pi_{\phi}}\left[-\alpha \log \pi_{\phi}\left(\mathbf{a} \mid \mathbf{s}\right)-\alpha \overline{\mathcal{H}}\right], \label{eq:sac_temp_loss}
\end{align}
where $\overline{\mathcal{H}}$ is the target entropy.

\subsection{Uncertainty punished Q-learning} \label{rac:main_method}
Uncertainty punished Q-learning(UPQ) is a variant of soft Bellman residual\eqref{eq:sac_critic_tot}. The idea is to maintain an ensemble of $N$ soft Q-functions $Q_{\theta_{i}}(\mathbf{s}, \mathbf{a})$, where $\theta_{i}$ denote the parameters of the $i-th$ soft Q-function, which are initialized randomly and independently for inducing an initial diversity in the models~\cite{osband2016deep}, but updated with the same target.

Given a transition $\tau_t$, UPQ consider following uncertainty punished target $y$:
\begin{align}
  & y=r_t+\gamma \mathbb{E}_{\mathbf{a'}\sim \pi_\phi}\left[\bar {Q}_{\bar \theta}(\mathbf{s'}, \mathbf{a'})-\beta \hat{s}({Q}_{\bar \theta}(\mathbf{s'}, \mathbf{a'}))-\alpha \log \pi_{\phi}\left(\mathbf{a'} \mid \mathbf{s'}\right)\right], \label{eq:rac_critic_target}
  \\
  & \bar{Q}_{\bar \theta}(\mathbf{s'}, \mathbf{a'}) = \frac{1}{N} \sum_{i=1}^{N}Q_{\bar \theta_{i}}(\mathbf{s'}, \mathbf{a'}), \label{eq:rac_critic_mean}
  \\
  & \hat{s}({Q}_{\bar \theta}(\mathbf{s'}, \mathbf{a'})) = \sqrt{\frac{1}{N-1} \sum_{i=1}^{N}\left[Q_{\bar \theta_{i}}(\mathbf{s'}, \mathbf{a'}) -\bar{Q}_{\bar \theta}(\mathbf{s'}, \mathbf{a'})\right]^{2}}, \label{eq:rac_critic_std}
\end{align}
where $\bar{Q}_{\bar \theta}(\mathbf{s}, \mathbf{a})$ is the sample mean of target Q-functions, $\hat{s}({Q}_{\bar \theta}(\mathbf{s}, \mathbf{a}))$ is the sample standard deviation of target Q-functions with bessel's correction~\cite{warwick1975sample}. UPQ uses $\hat{s}({Q}_{\bar \theta}(\mathbf{s}, \mathbf{a}))$ as uncertainty estimation to punish value estimation. $\beta \ge 0$ is the weighting of the punishment. Note
that we do not propagate gradient through the uncertainty $\hat{s}({Q}_{\bar \theta}(\mathbf{s}, \mathbf{a}))$.

We write $Q_{\mathbf{s} \mathbf{a}}^{i}$ instead of $Q_{\theta_{i}}(\mathbf{s}, \mathbf{a})$, $Q_{\mathbf{s'} \mathbf{a'}}^{i}$ instead of $Q_{\theta_{i}}(\mathbf{s'}, \mathbf{a'})$ for compactness. Assuming each Q-function has random approximation error $e_{\mathbf{s} \mathbf{a}}^{i}$~\cite{thrun1993issues, lan2020maxmin, chen2021randomized} which is a random variable belonging to some distribution
\begin{align}
  & Q_{\mathbf{s} \mathbf{a}}^{i}=Q_{\mathbf{s} \mathbf{a}}^{*}+e_{\mathbf{s} \mathbf{a}}^{i}, \label{eq:Q_bias}
\end{align}
where $Q_{\mathbf{s} \mathbf{a}}^{*}$ is the ground truth of Q-functions. $M$ is the number of actions applicable at state $\mathbf{s'}$. Define the estimation bias $Z_{M N}$ for a transition $\tau_t$ to be
\begin{align}
Z_{M N} & \stackrel{\text { def }}{=}\left[r+\gamma \max _{\mathbf{a}^{\prime}} (Q_{\mathbf{s}^{\prime} \mathbf{a}^{\prime}}^{mean}-\beta Q_{\mathbf{s}^{\prime} \mathbf{a}^{\prime}}^{std})\right]-\left(r+\gamma \max _{\mathbf{a}^{\prime}} Q_{\mathbf{s}^{\prime} \mathbf{a}^{\prime}}^{*}\right)
\\
&=\gamma\left[\max _{\mathbf{a}^{\prime}} (Q_{\mathbf{s}^{\prime} \mathbf{a}^{\prime}}^{mean}-\beta Q_{\mathbf{s}^{\prime} \mathbf{a}^{\prime}}^{std})-\max _{\mathbf{a}^{\prime}} Q_{\mathbf{s}^{\prime} \mathbf{a}^{\prime}}^{*}\right],
\end{align}
where 
\begin{align}
Q_{\mathbf{s}^{\prime} \mathbf{a}^{\prime}}^{mean} &\approx \frac{1}{N} \sum_{i=1}^{N}Q_{\mathbf{s'} \mathbf{a'}}^{i}= \frac{1}{N} \sum_{i=1}^{N} (Q_{\mathbf{s'} \mathbf{a'}}^{*}+e_{\mathbf{s'} \mathbf{a'}}^{i})= Q_{\mathbf{s'} \mathbf{a'}}^{*}+\frac{1}{N} \sum_{i=1}^{N} e_{\mathbf{s'} \mathbf{a'}}^{i}=Q_{\mathbf{s'} \mathbf{a'}}^{*}+\bar {e}_{\mathbf{s'} \mathbf{a'}},
\\
Q_{\mathbf{s}^{\prime} \mathbf{a}^{\prime}}^{std} &\approx \sqrt{\frac{1}{N-1} \sum_{i=1}^{N}\left(Q_{\mathbf{s}^{\prime} \mathbf{a}^{\prime}}^{i} -Q_{\mathbf{s}^{\prime} \mathbf{a}^{\prime}}^{mean}\right)^{2}} = \sqrt{\frac{1}{N-1} \sum_{i=1}^{N}\left(Q_{\mathbf{s'} \mathbf{a'}}^{*}+e_{\mathbf{s'} \mathbf{a'}}^{i} - Q_{\mathbf{s'} \mathbf{a'}}^{*}+ \bar {e}_{\mathbf{s'} \mathbf{a'}}\right)^{2}}
\\
&= \sqrt{\frac{1}{N-1} \sum_{i=1}^{N}\left(e_{\mathbf{s'} \mathbf{a'}}^{i} - \bar {e}_{\mathbf{s'} \mathbf{a'}}\right)^{2}}=\hat{s}(e_{\mathbf{s'} \mathbf{a'}}).
\end{align}
Then 
\begin{align}
& Z_{MN} \approx \gamma\left[\max _{\mathbf{a}^{\prime}} (Q_{\mathbf{s'} \mathbf{a'}}^{*}+\bar {e}_{\mathbf{s'} \mathbf{a'}}-\beta \hat{s}(e_{\mathbf{s'} \mathbf{a'}}))-\max _{\mathbf{a}^{\prime}} Q_{\mathbf{s}^{\prime} \mathbf{a}^{\prime}}^{*}\right].
\end{align}
If one could choose $\beta = \frac{\bar {e}_{\mathbf{s'} \mathbf{a'}}}{\hat{s}(e_{\mathbf{s'} \mathbf{a'}})}$, $Q_{\mathbf{s} \mathbf{a}}^{i}$ will be resumed to $Q_{\mathbf{s} \mathbf{a}}^{*}$, then $Z_{MN}$ can be reduced to near 0. However, it's hard to adjust a suitable constant $\beta$ for various state-action pairs actually. We develop vanilla RAC which uses a constant $\beta$ in section~\ref{ablation:Vanilla RAC} to research this problem. 

{\bf Shifting smoothly between higher and lower bounds}. For $\beta$ = 0, the update is simple average Q-learning which causes overestimation bias~\cite{chen2021randomized}. As $\beta$ increasing, increasingly penalties $Q_{\mathbf{s}^{\prime} \mathbf{a}^{\prime}}^{std}$ decrease $E[Z_{MN}]$ gradually, and encourage Q-functions transit smoothly from higher-bounds to lower-bounds.

{\bf Stable target estimation}.
Standard deviation and mean of target Q-functions used in UPQ are not sensitive to function approximation errors resulting a stable target estimation.

\subsection{Realistic actor-critic agent} \label{sec:agent}
\begin{algorithm}[t]
\caption{RAC: SAC version} \label{alg:rac_sac}
\begin{algorithmic}[1]
\State Initialize actor network $\phi$, $N$ critic networks $\theta_i, i=1,\ldots,N$, temperature network $\psi$, empty replay buffer $\mathcal{B}$, target network $\bar{\theta_{i}}\longleftarrow \theta_{i}$, for $i=1,2,\ldots,N$, uniform distribution $U_{1}$ and $U_{2}$
\For{each iteration}
\State execute an action $a \sim \pi_{\phi}\left(\cdot \mid s, \beta \right), \beta \sim U_{2}$. Observe reward $r_t$, new state $s'$
\State Store transition tuple $\mathcal{B} \leftarrow \mathcal{B} \cup\left\{\left(s, a, r_{t}, s'\right)\right\}$
\For{$G$ updates}
\State Sample random minibatch  $\{\tau_j\}_{j=1}^B\sim\mathcal{B}$, $\{\beta_m\}_{m=1}^B\sim U_{1}$
\State Compute the Q target~\eqref{eq:rac_critic_target_bata}
\For{$i=1,\ldots,N$}
\State Update $\theta_i$ by minimize $\mathcal{L}^{\tt RAC}_{\tt critic}$~\eqref{eq:rac_critic_tot_beta}
\renewcommand\baselinestretch{1.2}\selectfont
\State Update target networks with $\bar{\theta_{i}} \leftarrow \rho \bar{\theta_{i}}+(1-\rho) \theta_{i}$
\EndFor 
\renewcommand\baselinestretch{1.1}\selectfont
\EndFor 
\State Update actor parameters $\phi$ by minimize $\mathcal{L}^{\tt RAC-SAC}_{\tt actor} $~\eqref{eq:rac_actor_tot}
\State Update temperature parameters $\psi$ by minimize $\mathcal{L}^{\tt RAC}_{\tt temp}$~\eqref{eq:rac_temp_loss}
\EndFor
\end{algorithmic}
\end{algorithm}
We now demonstrate how to use UPQ to incorporate various bounds of value approximations into a full agent that maintains diverse policies, each with a different under-overestimation trade-off. The pseudocode for RAC-SAC is shown in Algorithm~\ref{alg:rac_sac}.

RAC use a UVFA~\cite{schaul2015universal} to extend the critic and actor as $Q_{\theta_{i}}(\mathbf{s},\mathbf{a},\beta)$ and $\pi_{\phi}\left(\cdot \mid \mathbf{s'}, \beta \right)$, $U_{1}$ is a uniform traning distribution $\mathcal{U}[0, a]$, $a$ a is a positive real number, $\beta \sim U_{1}$ that generates various bounds of value approximations.

An independent temperature network $\alpha_{\psi}$ parameterized by $\psi$ is used to accurately adjust the temperature with respect to $\beta$, which can improve the performance of RAC. Then the objective~\eqref{eq:sac_temp_loss} becomes:
\begin{align}
  & \mathcal{L}^{\tt RAC}_{\tt temp} (\psi) = \mathbb{E}_{\mathbf{s} \sim \mathcal{B}, \mathbf{a} \sim \pi_{\phi}, \beta \sim U_{1}}\left[-\alpha_{\psi}(\beta) \log \pi_{\phi}\left(\mathbf{a} \mid \mathbf{s}, \beta \right)-\alpha_{\psi}(\beta) \overline{\mathcal{H}}\right]. \label{eq:rac_temp_loss}
\end{align}
The extended Q-ensemble use UPQ to simultaneously approximate a soft Q-function family:
\begin{align}
  & \mathcal{L}^{\tt RAC}_{\tt critic} (\theta_i) = \mathbb{E}_{\tau \sim \mathcal{B}, \beta \sim U_{1}} [\left(Q_{\theta_{i}}(\mathbf{s},\mathbf{a},\beta) - y \right)^2],\label{eq:rac_critic_tot_beta}
    \\
  & y = r+\gamma \mathbb{E}_{\mathbf{a'}\sim \pi_\phi} [\bar{Q}_{\bar \theta}(\mathbf{s'},\mathbf{a'},\beta)-\beta \hat{s}({Q}_{\bar \theta}(\mathbf{s'},\mathbf{a'},\beta))-\alpha_{\psi}(\beta) \log \pi_{\phi}(\mathbf{a'} \mid \mathbf{s'},\beta)], \label{eq:rac_critic_target_bata}
\end{align}
where $\bar{Q}_{\bar \theta}(\mathbf{s},\mathbf{a},\beta)$ is the sample mean of target Q-functions, $\hat{s}({Q}_{\bar \theta}(\mathbf{s},\mathbf{a},\beta))$ is the corrected sample standard deviation of target Q-functions.

The extended policy $\pi_{\phi}$ is updated by minimizing the following object:
\begin{align}
  & \mathcal{L}^{\tt RAC-SAC}_{\tt actor} (\phi) = \mathbb{E}_{\mathbf{s} \sim \mathcal{B}, \beta \sim U_{1}}\left[\mathbb{E}_{\mathbf{a} \sim \pi_{\phi}}\left[\alpha_{\psi}(\beta) \log \left(\pi_{\phi}\left(\mathbf{a} \mid \mathbf{s}, \beta\right)\right)-\bar Q_{\theta}\left(\mathbf{a}, \mathbf{s}, \beta\right)\right]\right], \label{eq:rac_actor_tot}
\end{align}
where $\bar Q_{\theta}\left(\mathbf{a}, \mathbf{s}, \beta\right)$ is the sample mean of Q-functions.

Note that, we find that apply different samples, which are generated by binary masks from the Bernoulli distribution~\cite{lee2020sunrise, osband2016deep}, to train each Q-function won’t improve RAC performance in our experiments, therefore RAC does not apply this method. 

{\bf RAC circumvents direct adjustment of $\beta$}. RAC leaners with a distribution of $\beta$ instead of a constant $\beta$. one could evaluate the policy family to find the best $\beta$. We employ a discrete number $H$ of values $\left\{\beta_{i}\right\}_{i=1}^{H}$ (see details in Appendix ~\ref{appendix:evaluationmethod}) to implement a distributed evaluation for computational efficiency, and apply the max strategy to get best $\beta$.

{\bf Optimistic exploration}.
When interacting with the environment, we propose to sample $\beta$ uniformly from a uniform explore distribution $U_{2} = \mathcal{U}[0, b]$,where $b < a$ is a positive real number, to get optimistic exploratory behaviors to avoid pessimistic underexploration~\cite{ciosek2019better}. The diversified policies with respect to different $\beta$ generate varied action sequences to visit unseen state-action pairs following the principle of optimism in the face of uncertainty~\cite{ciosek2019better, lee2020sunrise, chen2017ucb}.

{\bf Sample efficiency}.
RAC improves sample efficiency from two aspects: (1) Larger UTD ratio $G$ improves samples utilization. (2) Learning smoothly and diverse policies in the same network build a powerful representation and set of skills that can be quickly transferred to the expected policy. And we find that a smaller replay buffer capacity slightly improves the sample efficiency of RAC in section~\ref{ablation:smallercapacity}. 


\section{Experiments} \label{sec:experiments}
We designed our experiments to answer the following questions: 
\begin{itemize} [leftmargin=8mm]
\setlength\itemsep{0.1em}
  \item Can Realistic Actor-Critic outperform state-of-the-art algorithms in continuous control tasks?
  \item Can uncertainty punished Q-learning(UPQ) improve the quality of value approximation?
  \item What is the contribution of each technique in Realistic Actor-Critic?
\end{itemize}

\begin{figure*}[ht]
  \centering
  \includegraphics[width=\textwidth]{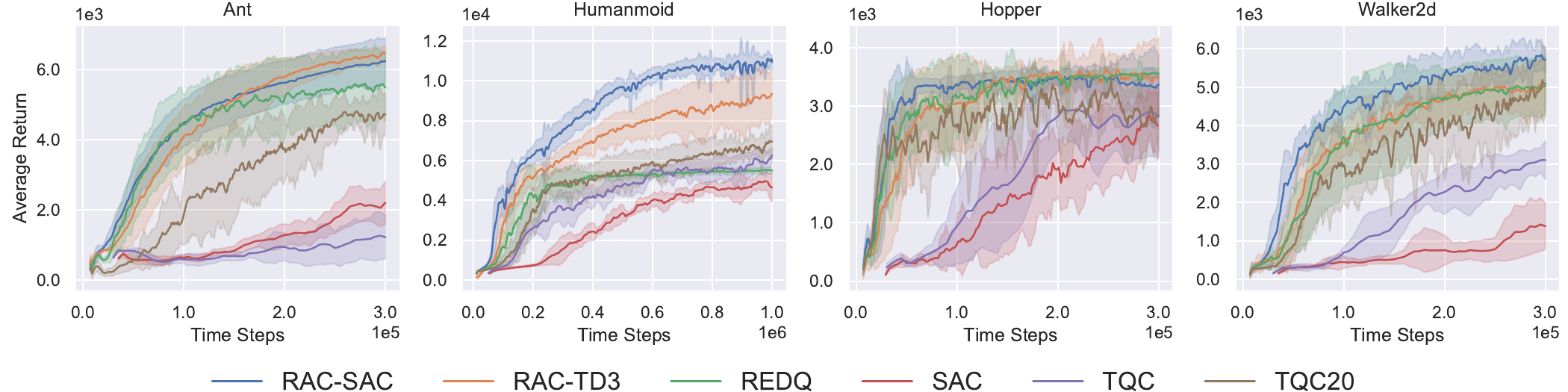}
  \caption{The learning curves on 4 Mujoco environments. The horizontal axis indicates number of time steps. The vertical axis indicates the average undiscounted return. The shaded areas denote one standard deviation.}
  \label{fig-performance}
\end{figure*}

\begin{figure*}[ht]
  \centering
  \includegraphics[width=\textwidth]{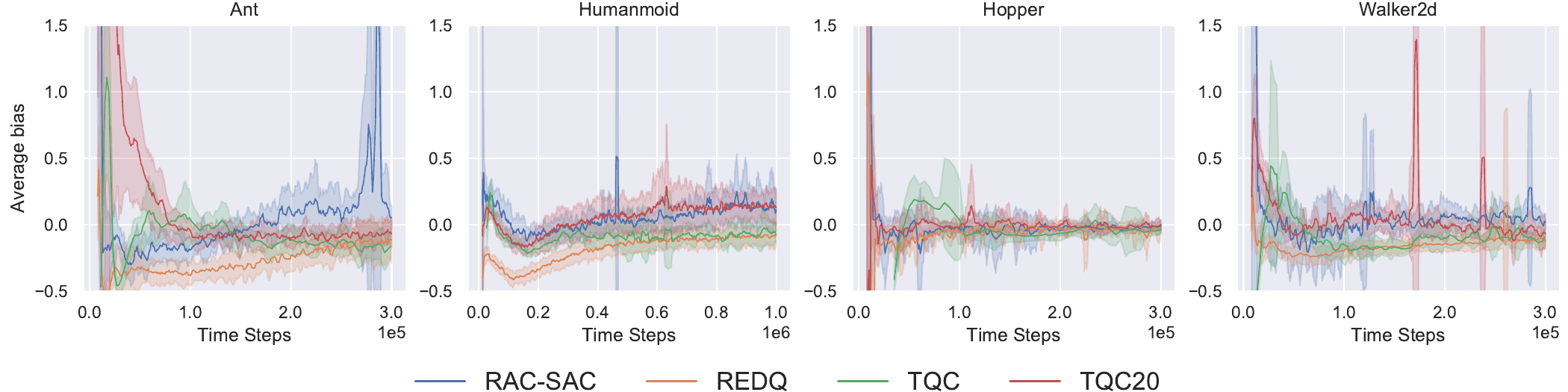}
  \caption{Mean of normalized Q bias of RAC-SAC, REDQ, TQC and TQC20 for mujoco environments.}
  \label{fig-norm-mean}
\end{figure*}

\begin{figure*}[ht]
  \centering
  \includegraphics[width=\textwidth]{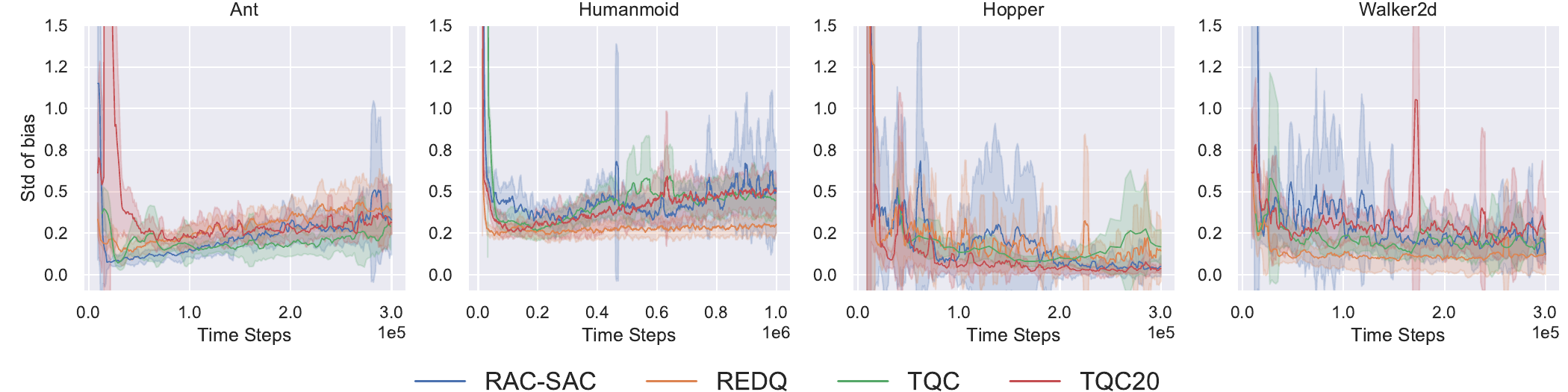}
  \caption{Std of normalized Q bias of RAC-SAC, REDQ, TQC and TQC20 for mujoco environments.}
  \label{fig-norm-std}
\end{figure*}

\subsection{Setups}
We implement RAC with SAC and TD3 as RAC-SAC and RAC-TD3(see more details in Appendix~\ref{appendix:hyperparameters}). We compare to state-of-the-art algorithms: SAC~\cite{haarnoja2018soft}, TD3~\cite{fujimoto2018addressing}, MBPO~\cite{janner2019trust}, REDQ~\cite{chen2021randomized} and  TQC~\cite{kuznetsov2020controlling} on 4 challenging continuous control tasks (Walker2d, HalfCheetah, Ant and Humanoid) from MuJoCo environments~\cite{todorov2012mujoco} in the OpenAI gym benchmark~\cite{brockman2016openai}. We alse implement TQC20 which is a varient of TQC with UTD $G=20$ for a fair comparison.

The time steps for training instances on Walker2d, Hopper, and Ant are $3\times 10^{5}$, and $1\times 10^{6}$ for Humanoid. All algorithms explore with a stochastic policy but use a deterministic policy for evaluation which is similar to those in SAC. We report the mean and standard deviation across 8 seeds. To analysis the value approximation quality, we calculate the mean and std of normalized values bias as main analysis indicators following REDQ~\cite{chen2021randomized}(described in Appendix~\ref{appendix:metric}). The average bias lets us know whether $Q_\theta$ is overestimated or underestimated, while std measures whether $Q_\theta$ is overfitting.

\subsection{Comparative evaluation}

\begin{table} [t]
\centering
\resizebox{\textwidth}{!}{
\begin{tabular}{l|ccccc|ccc}
\toprule
& RAC-SAC
& RAC-TD3
& REDQ
& MBPO
& TQC20
& TD3
& SAC
& TQC \\
\midrule
Humanmoid
& {\begin{tabular}[c]{@{}c@{}} {\bf 11107\scriptsize{$\pm$475}} \end{tabular}}
& {\begin{tabular}[c]{@{}c@{}} 9321\scriptsize{$\pm$1126} \end{tabular}} 
& {\begin{tabular}[c]{@{}c@{}} 5504\scriptsize{$\pm$120} \end{tabular}} 
& {\begin{tabular}[c]{@{}c@{}} 5162\scriptsize{$\pm$350} \end{tabular}} 
& {\begin{tabular}[c]{@{}c@{}} 7053\scriptsize{$\pm$857} \end{tabular}} 
& {\begin{tabular}[c]{@{}c@{}} 7014\scriptsize{$\pm$643} \end{tabular}}
& {\begin{tabular}[c]{@{}c@{}} 7681\scriptsize{$\pm$1118} \end{tabular}} 
& {\begin{tabular}[c]{@{}c@{}} 10731\scriptsize{$\pm$1296} \end{tabular}} \\
Ant
& {\begin{tabular}[c]{@{}c@{}} 6283\scriptsize{$\pm$549} \end{tabular}}
& {\begin{tabular}[c]{@{}c@{}} 6470\scriptsize{$\pm$165} \end{tabular}} 
& {\begin{tabular}[c]{@{}c@{}} 5475\scriptsize{$\pm$890} \end{tabular}} 
& {\begin{tabular}[c]{@{}c@{}} 5281\scriptsize{$\pm$699} \end{tabular}} 
& {\begin{tabular}[c]{@{}c@{}} 4722\scriptsize{$\pm$567} \end{tabular}} 
& {\begin{tabular}[c]{@{}c@{}} {\bf 6796\scriptsize{$\pm$277}} \end{tabular}}
& {\begin{tabular}[c]{@{}c@{}} 6433\scriptsize{$\pm$332} \end{tabular}}
& {\begin{tabular}[c]{@{}c@{}} 6402\scriptsize{$\pm$1371} \end{tabular}}\\
Walker
& {\begin{tabular}[c]{@{}c@{}} {\bf 5860\scriptsize{$\pm$440}} \end{tabular}}
& {\begin{tabular}[c]{@{}c@{}} 5114\scriptsize{$\pm$489} \end{tabular}} 
& {\begin{tabular}[c]{@{}c@{}} 5034\scriptsize{$\pm$711} \end{tabular}} 
& {\begin{tabular}[c]{@{}c@{}} 4864\scriptsize{$\pm$488} \end{tabular}} 
& {\begin{tabular}[c]{@{}c@{}} 5109\scriptsize{$\pm$696} \end{tabular}} 
& {\begin{tabular}[c]{@{}c@{}} 4419\scriptsize{$\pm$1682} \end{tabular}}
& {\begin{tabular}[c]{@{}c@{}} 5249\scriptsize{$\pm$554} \end{tabular}}
& {\begin{tabular}[c]{@{}c@{}} 5821\scriptsize{$\pm$457} \end{tabular}}\\
Hopper
& {\begin{tabular}[c]{@{}c@{}} 3421\scriptsize{$\pm$483} \end{tabular}} 
& {\begin{tabular}[c]{@{}c@{}} 3495\scriptsize{$\pm$672} \end{tabular}} 
& {\begin{tabular}[c]{@{}c@{}} {\bf 3563\scriptsize{$\pm$94}} \end{tabular}} 
& {\begin{tabular}[c]{@{}c@{}} 3280\scriptsize{$\pm$455} \end{tabular}} 
& {\begin{tabular}[c]{@{}c@{}} 3208\scriptsize{$\pm$538} \end{tabular}} 
& {\begin{tabular}[c]{@{}c@{}} 3433\scriptsize{$\pm$321} \end{tabular}}
& {\begin{tabular}[c]{@{}c@{}} 2815\scriptsize{$\pm$585} \end{tabular}}
& {\begin{tabular}[c]{@{}c@{}} 3011\scriptsize{$\pm$866} \end{tabular}}\\

\bottomrule
\end{tabular}}
\vspace{0.1in}
\caption{Performance on OpenAI Gym. The best results are indicated in bold. Performances of SAC, TD3 and TQC are obtained with 10x samples at $6\times10^6$ timesteps for Humanmoid and $3\times 10^6$ timesteps for other environments.}
\label{tbl:Comparative_Evaluation}
\end{table}

\begin{table} [t]
\centering
\resizebox{\textwidth}{!}{
\begin{tabular}{l|ccccc|cccc}
\toprule
& RAC-SAC
& REDQ
& MBPO
& TQC
& TQC20
& REDQ/RAC-SAC
& MBPO/RAC-SAC
& TQC/RAC-SAC
& TQC20/RAC-SAC\\
\midrule
Humanmoid at $2000$
& {\begin{tabular}[c]{@{}c@{}} 63K \end{tabular}}
& {\begin{tabular}[c]{@{}c@{}} 109K \end{tabular}} 
& {\begin{tabular}[c]{@{}c@{}} 154K \end{tabular}} 
& {\begin{tabular}[c]{@{}c@{}} 145K \end{tabular}} 
& {\begin{tabular}[c]{@{}c@{}} 147K \end{tabular}} 
& {\begin{tabular}[c]{@{}c@{}} 1.73 \end{tabular}}
& {\begin{tabular}[c]{@{}c@{}} 2.44 \end{tabular}}
& {\begin{tabular}[c]{@{}c@{}} 2.30 \end{tabular}} 
& {\begin{tabular}[c]{@{}c@{}} 2.33 \end{tabular}}\\
Humanmoid at $5000$
& {\begin{tabular}[c]{@{}c@{}} 134K \end{tabular}}
& {\begin{tabular}[c]{@{}c@{}} 250K \end{tabular}}
& {\begin{tabular}[c]{@{}c@{}} 295K \end{tabular}} 
& {\begin{tabular}[c]{@{}c@{}} 445K \end{tabular}} 
& {\begin{tabular}[c]{@{}c@{}} 258K \end{tabular}} 
& {\begin{tabular}[c]{@{}c@{}} 1.87 \end{tabular}}
& {\begin{tabular}[c]{@{}c@{}} 2.20 \end{tabular}}
& {\begin{tabular}[c]{@{}c@{}} 3.32 \end{tabular}} 
& {\begin{tabular}[c]{@{}c@{}} 1.93 \end{tabular}}\\
Humanmoid at $10000$
& {\begin{tabular}[c]{@{}c@{}} 552K\ \end{tabular}}
& {\begin{tabular}[c]{@{}c@{}} - \end{tabular}} 
& {\begin{tabular}[c]{@{}c@{}} - \end{tabular}} 
& {\begin{tabular}[c]{@{}c@{}} 3260K \end{tabular}} 
& {\begin{tabular}[c]{@{}c@{}} - \end{tabular}} 
& {\begin{tabular}[c]{@{}c@{}} - \end{tabular}}
& {\begin{tabular}[c]{@{}c@{}} - \end{tabular}}
& {\begin{tabular}[c]{@{}c@{}} 5.91 \end{tabular}} 
& {\begin{tabular}[c]{@{}c@{}} - \end{tabular}}\\
Ant at $1000$
& {\begin{tabular}[c]{@{}c@{}} 21K \end{tabular}}
& {\begin{tabular}[c]{@{}c@{}} 28K \end{tabular}} 
& {\begin{tabular}[c]{@{}c@{}} 62K \end{tabular}} 
& {\begin{tabular}[c]{@{}c@{}} 185K \end{tabular}} 
& {\begin{tabular}[c]{@{}c@{}} 42K \end{tabular}} 
& {\begin{tabular}[c]{@{}c@{}} 1.33 \end{tabular}}
& {\begin{tabular}[c]{@{}c@{}} 2.95 \end{tabular}}
& {\begin{tabular}[c]{@{}c@{}} 8.81 \end{tabular}}
& {\begin{tabular}[c]{@{}c@{}} 2.00 \end{tabular}}\\
Ant at $3000$
& {\begin{tabular}[c]{@{}c@{}} 56K \end{tabular}}
& {\begin{tabular}[c]{@{}c@{}} 56K \end{tabular}} 
& {\begin{tabular}[c]{@{}c@{}} 152K \end{tabular}} 
& {\begin{tabular}[c]{@{}c@{}} 940K \end{tabular}} 
& {\begin{tabular}[c]{@{}c@{}} 79K \end{tabular}} 
& {\begin{tabular}[c]{@{}c@{}} 1.00\end{tabular}}
& {\begin{tabular}[c]{@{}c@{}} 2.71 \end{tabular}}
& {\begin{tabular}[c]{@{}c@{}} 16.79 \end{tabular}} 
& {\begin{tabular}[c]{@{}c@{}} 1.41 \end{tabular}}\\
Ant at $6000$
& {\begin{tabular}[c]{@{}c@{}} 248K \end{tabular}}
& {\begin{tabular}[c]{@{}c@{}} - \end{tabular}} 
& {\begin{tabular}[c]{@{}c@{}} - \end{tabular}} 
& {\begin{tabular}[c]{@{}c@{}} 3055K \end{tabular}} 
& {\begin{tabular}[c]{@{}c@{}} - \end{tabular}} 
& {\begin{tabular}[c]{@{}c@{}} -\end{tabular}}
& {\begin{tabular}[c]{@{}c@{}} - \end{tabular}}
& {\begin{tabular}[c]{@{}c@{}} 12.31 \end{tabular}} 
& {\begin{tabular}[c]{@{}c@{}} - \end{tabular}}\\
Walker at $1000$
& {\begin{tabular}[c]{@{}c@{}} 27K \end{tabular}}
& {\begin{tabular}[c]{@{}c@{}} 42K \end{tabular}} 
& {\begin{tabular}[c]{@{}c@{}} 54K \end{tabular}} 
& {\begin{tabular}[c]{@{}c@{}} 110K \end{tabular}} 
& {\begin{tabular}[c]{@{}c@{}} 50K \end{tabular}} 
& {\begin{tabular}[c]{@{}c@{}} 1.56\end{tabular}}
& {\begin{tabular}[c]{@{}c@{}} 2.00 \end{tabular}}
& {\begin{tabular}[c]{@{}c@{}} 4.07 \end{tabular}} 
& {\begin{tabular}[c]{@{}c@{}} 1.85 \end{tabular}}\\
Walker at $3000$
& {\begin{tabular}[c]{@{}c@{}} 53K \end{tabular}}
& {\begin{tabular}[c]{@{}c@{}} 79K \end{tabular}} 
& {\begin{tabular}[c]{@{}c@{}} 86K \end{tabular}} 
& {\begin{tabular}[c]{@{}c@{}} 270K \end{tabular}} 
& {\begin{tabular}[c]{@{}c@{}} 89K \end{tabular}} 
& {\begin{tabular}[c]{@{}c@{}} 1.49\end{tabular}}
& {\begin{tabular}[c]{@{}c@{}} 1.62 \end{tabular}}
& {\begin{tabular}[c]{@{}c@{}} 10.75 \end{tabular}} 
& {\begin{tabular}[c]{@{}c@{}} 1.68 \end{tabular}}\\
Walker at $5000$
& {\begin{tabular}[c]{@{}c@{}} 147K \end{tabular}}
& {\begin{tabular}[c]{@{}c@{}} 272K \end{tabular}} 
& {\begin{tabular}[c]{@{}c@{}} - \end{tabular}} 
& {\begin{tabular}[c]{@{}c@{}} 960K \end{tabular}} 
& {\begin{tabular}[c]{@{}c@{}} 270K \end{tabular}} 
& {\begin{tabular}[c]{@{}c@{}} 1.85\end{tabular}}
& {\begin{tabular}[c]{@{}c@{}} - \end{tabular}}
& {\begin{tabular}[c]{@{}c@{}} 6.53 \end{tabular}} 
& {\begin{tabular}[c]{@{}c@{}} 1.84 \end{tabular}}\\
\bottomrule
\end{tabular}}
\vspace{0.1in}
\caption{Sample efficiency (SE)~\cite{chen2021randomized, dorner2021measuring} is measure by the ratio of the number of samples collected when RAC and some algorithm reach the specified performance. The last 4 rows show how many
times REDQ are more sample efficient than other algorithms in reaching that performance. Hopper is not in the comparison object as the performance of algorithms is almost indistinguishable.}
\label{tbl:Sample_efficience}
\end{table}

{\bf OpenAI Gym}. Figure \ref{fig-performance} and Table~\ref{tbl:Comparative_Evaluation} shows learning curves and performance comparison. RAC consistently improves the performance of SAC and TD3 across all environments and performs consistently better than other algorithms especially in Humanmoid. Table~\ref{tbl:Comparative_Evaluation} shows that RAC-SAC can outperform SAC’s performance with about one-tenth of the samples.
It is seen that RAC yields a much smaller variance than SAC and TQC which implies that the optimistic exploration helps the agents escape out of bad local optima.

{\bf Sample efficiency comparison}. Results in table \ref{tbl:Sample_efficience} show that the sampling efficiency of RAC exceeds other algorithms. Compared with TQC, RAC-SAC reaches 3000 and 6000 for Ant with 16.79x and 12.31x sample efficiency. RAC-SAC performs 1.5x better than REDQ half-way through training and 1.8x better at the end of
training in Walker and Huamnmoid.
The sample efficiency of TQC20 is also significantly improved compared to TQC which show that the UTD ratio is indeed a key factor to sample efficiency.

{\bf Value approximation analysis}. Figure \ref{fig-norm-mean} and \ref{fig-norm-std} presents std and mean of normalized Q bias. In Ant and humanmoid, RAC-SAC quickly suppresses overestimation at the beginning of training and reduces the std to a lower level. Different with REDQ which always keeps negative Q bias, RAC-SAC slowly moves Q bias towards from underestimation to overestimation without injuring performance. 
This abnormal phenomenon indicates that overestimation can still effectively improve the performance of agents in some situation which is consistent with ~\citet{lan2020maxmin}'s view.

\begin{figure*}[ht]
  \centering
  \includegraphics[width=\textwidth]{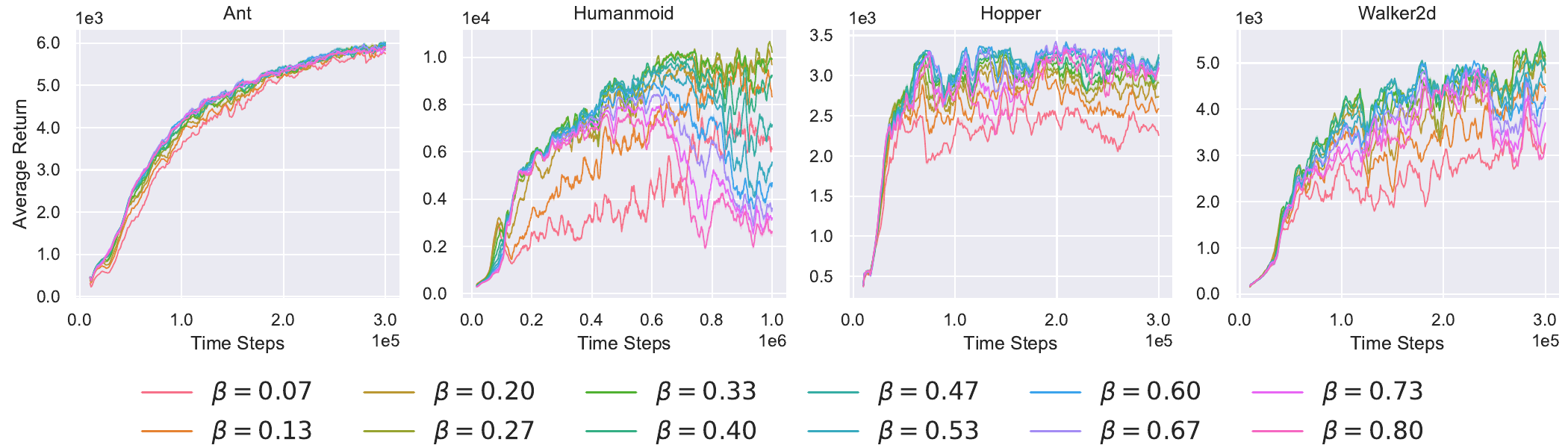}
  \caption{Performance of various value confidence bounds with respect to different $\beta$ during training for RAC-SAC.}
  \label{Performance_Visualisations}
\end{figure*}

\begin{figure*}[ht]
  \centering
  \includegraphics[width=\textwidth]{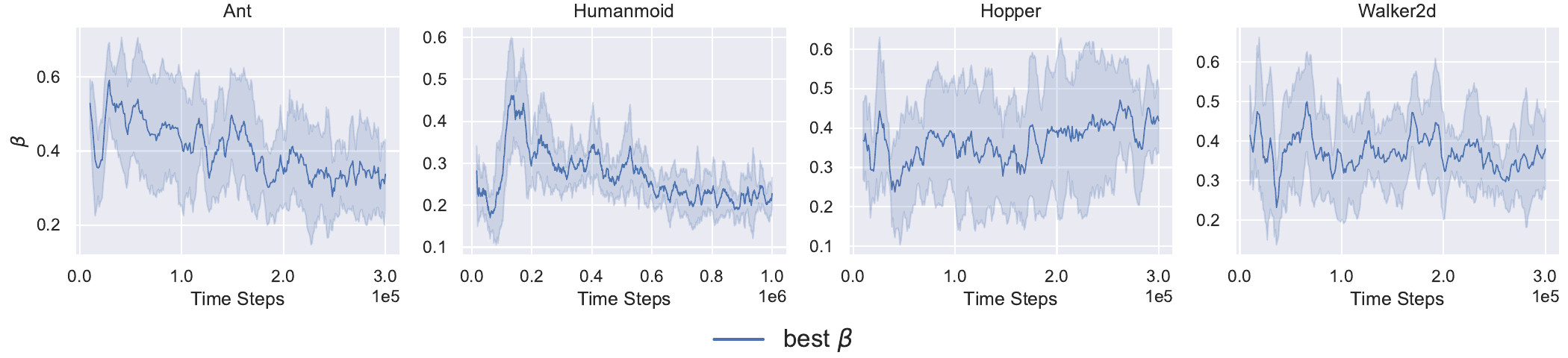}
  \caption{Visualisations of learned best $\beta$ for RAC-SAC. RAC can always auto find the best $\beta$ for different environment to balance between under- and overestimation with simple max strategy.}
  \label{best_beta_Visualisations}
\end{figure*}

{\bf Performance of different confidence bounds}. Results in Figure \ref{Performance_Visualisations} show that Ant is not sensitive to value confidence bounds, while small changes in $\beta$ will have a huge impact on the performance of RAC-SAC at Humanmoid, Walker2d and Hopper. Figure \ref{best_beta_Visualisations} shows the best $\beta$ for different environments and time steps can be completely different, using a constant $\beta$ is unreasonable.

\subsection{Variants of RAC}

\begin{figure*}[ht]
  \centering
  \includegraphics[width=\textwidth]{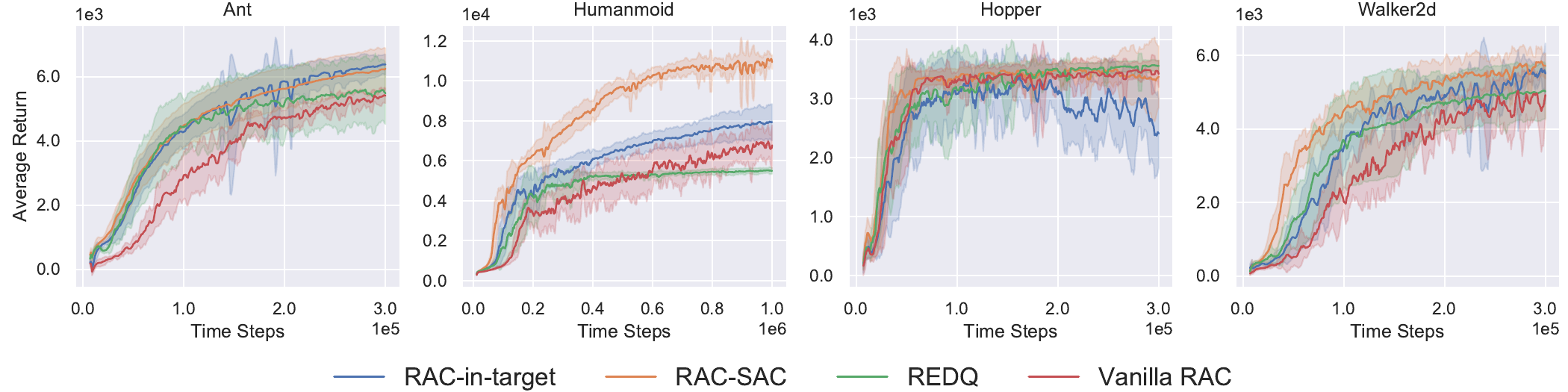}
  \caption{Performance of RAC and its variants.}
  \label{RACs_main_performance}
\end{figure*}

To study the role of uncertainty punished Q-learning(UPQ), we implement two variants of RAC:

{\bf Vanilla RAC}\label{ablation:Vanilla RAC}. Vanilla RAC uses UPQ with a constant $\beta$ to replace in-target minimization in REDQ resulting in simple vanilla RAC. The pseudocode for vanilla RAC is shown in Algorithm~\ref{alg:vanilla_rac}. In such a way, we can test whether UPQ improves performance compared to in-target minimization. 

{\bf RAC with in-target minimization}. To study the contribution of UPQ to RAC, we implement a variant of RAC-SAC which uses in-target minimization instead of UPQ to train the Q-ensemble. The pseudocode for vanilla RAC is shown in Algorithm~\ref{alg:RAC_in_target_minimization}.

The implement details can be found in Appendix~\ref{appendix:vanilla_rac} and~\ref{appendix:RAC_in_target_minimization}. Results in figure \ref{RACs_main_performance} show that the performance of vanilla RAC is just as good as REDQ for most of the training. Compared with RAC-SAC, lower performance of vanilla RAC indicates other components of RAC-SAC are critical to performance. 

In Humanmoid, the performance of RAC with in-target minimization is significantly lower than that of RAC-SAC which means UPQ is essential for RAC-SAC.

\subsection{Ablation study}

\begin{figure*}[ht]
  \centering
  \includegraphics[width=\textwidth]{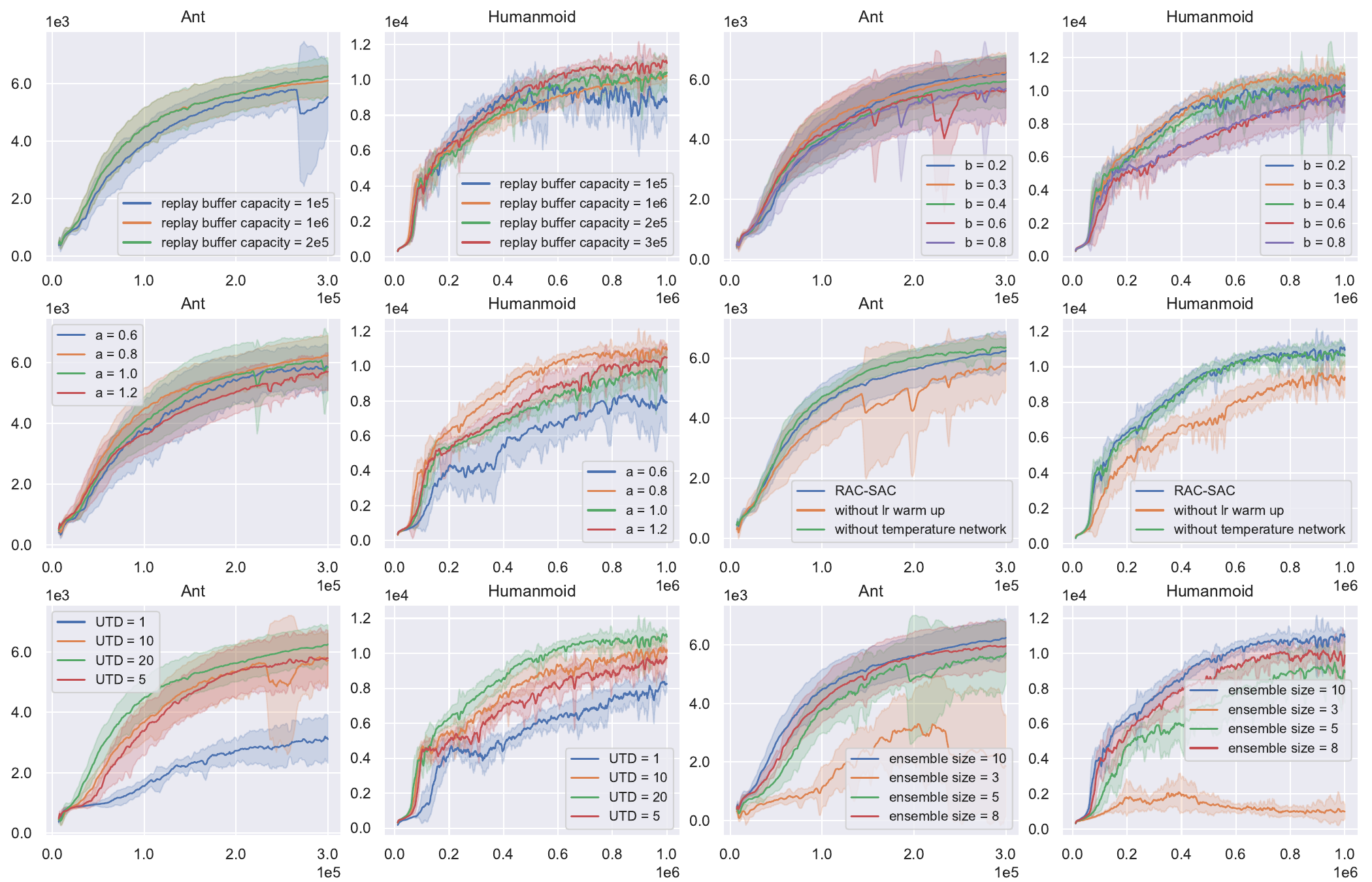}
  \caption{Results of ablations for RAC-SAC. Results are averaged over 8 seeds, ± std is shaded. The horizontal axis indicates number of time steps. The vertical axis indicates the average undiscounted return. }
  \label{fig-ablation}
\end{figure*}


{\bf Smaller replay buffer capacity}\label{ablation:smallercapacity}. In this experiment we vary the replay buffer capacity in $\{5\times10^4, 10^5,  2\times10^5, 3\times10^5, 10^6\}$. The results in figure \ref{fig-ablation} show that RAC-SAC can benefit from a smaller capacity, but will be hurt when the capacity is excessively small. 

{\bf Optimistic exploration}.  We vary RAC-SAC with $b\in\{0.2, 0.3, 0.4, 0.6, 0.8\}$ for exploration distribution $U_{2}[0, b]$. The results in figure \ref{fig-ablation} show that 
exploratory policies become more conservative with $b$ increasing, and the performance of RAC-SAC gradually declines. The increasing standard deviation means that more and more agents fall into suboptimal policies. But if $b$ is too small, the lack of exploratory diversity will cause performance degradation.

{\bf Training distribution}. We vary RAC-SAC with $a\in\{0.4, 0.6, 0.8, 1.0, 1.2\}$ for training distribution $U_{1}[0, a]$. The results in figure \ref{fig-ablation} show that 
the average bias becomes stable with $a$ increasing. The $Q_\theta$ estimates will suffer a large negative bias when $a$ is too large, which make policies too conservative.

{\bf Independent temperature network}. We implement a variant of RAC-SAC whose temperature is adjusted following rules~\eqref{eq:sac_temp_loss} instead of using a temperature network. Figure \ref{fig-ablation} shows that an independent temperature network has a relatively small impact on RAC-SAC performance. Figure \ref{temp_Visualisations} in Appendix~\ref{appendix:temp_Visualisations} shows temperatures learned by the temperature network belong to different $\beta$ at some time steps.

{\bf Update-To-Data (UTD) ratio}. Figure \ref{fig-ablation} shows performance of RAC-SAC under UTD ratio values $G\in\{1, 5, 10, 20\}$. The results suggest higher UTD values significantly improve the sample efficiency of RAC-SAC, with $G=20$ giving the best result.

{\bf Ensemble size}. We vary ensemble size $N\in\{3, 5, 8, 10\}$. The results in figure \ref{fig-ablation} suggest a larger ensemble size can stabilize average bias and decrease std of bias bringing about better performance.


{\bf Learning rate warm up}. RAC-SAC applies a linear learning rate warm-up strategy (see details in Appendix~\ref{appendix:hyperparameters}) to stable early learning. The results in figure \ref{fig-ablation} demonstrate that the early learning of RAC-SAC will be unstable without the learning rate warm-up.

\section{Conclusion}

In this paper, we present RAC to balance value under- and overestimation by involving policies with different value confidence-bonds. RAC is a simple ensemble method, which can be used to improve off-policy reinforcement learning algorithms. Experiments show advantageous properties of RAC: low value approximation error and brilliant sample efficiency. Results on continuous control benchmarks suggest that RAC consistently improves performances of existing off-policy RL algorithms, such as SAC and TD3.

Our results suggest that directly incorporate uncertainty to value functions and learning a powerful policy family can provide a promising avenue for improved sample efficiency and performance, and further exploration of ensemble methods, including high-level policies or more rich policy classes is an exciting avenue for future work.

\section*{Acknowledgements}
The authors gratefully acknowledge the financial support from National Natural Science Foundation of China (Grant No. 51779059), National Natural Science Foundation of Heilongjiang Province (Grant No.YQ2020E028).

\bibliography{reference}
\bibliographystyle{ref_bst}

\input{appendix}

\end{document}

%% file: appendix.tex
\clearpage
\appendix
\begin{center}{\bf {\LARGE Appendix:}}
\end{center}
\begin{center}{\bf {\Large \mytitle}}
\end{center}

\section{Experiments Setups and methodology} \label{appendix:setup}

\subsection{Reproducing the Baselines}
The baseline algorithms are REDQ~\cite{chen2021randomized}, MBPO~\cite{janner2019trust}, SAC~\cite{haarnoja2018soft}, TD3~\cite{fujimoto2018addressing} and TQC~\cite{kuznetsov2020controlling}. All hyper-parameters we used for evaluation are the same as those in the original papers. For MBPO\footnote{\url{https://github.com/JannerM/mbpo}},  REDQ\footnote{\url{https://github.com/watchernyu/REDQ}}, TD3\footnote{\url{https://github.com/sfujim/TD3}} and TQC\footnote{\url{https://github.com/SamsungLabs/tqc_pytorch}}, we use the authors’s code. For SAC, we implement it following~\citet{haarnoja2018soft}, and results we obtained are similar to previously reported results.

\subsection{Evaluation method} \label{appendix:evaluationmethod}
For all training instances, the policies are evaluated every $R_{eval} = 10^{3}$ time steps. At each evaluation phase, the agent fixes its policy, deterministically interacts with the same environment separate to obtain $10$ episodic rewards. The mean and standard deviation of these $10$ episodic rewards is the performance metrics of the agent at the evaluation phase. 

In the case of RAC, we employ a discrete number $H$ of values $\left\{\beta_{i}\right\}_{i=1}^{H}$ to get $H$ policies:
\begin{align}
  & \beta_{i}=b/H \cdot i, i=1,\ldots,H. \label{eq:beta}
\end{align}
Each of the $H$ policies fixes its policy as the one at the evaluation phase and deterministically interacts with the environment with the fixed policy to obtain $10$ episodic rewards. The $10$ episodic rewards are averaged for each policy, and then the maximum of the $10$-episode-average rewards of the $H$ policies is taken as the performance at that evaluation phase. 

We performed this operation for $8$ different random seeds used in the computational packages(numpy, pytorch) and environments(openai gym), and the mean and standard deviation of the learning curve are obtained from these $8$ simulations. 

For all experiments, our learning curves show the total undiscounted return.

\subsection{The normalized values bias estimation} \label{appendix:metric}
Given a state-action pair, the normalized values bias is defined as:
\begin{align}
  & \left(\bar Q_{\theta }(\mathbf{s},\mathbf{a}) - Q^{\pi}(\mathbf{s}, \mathbf{a})\right) /\left|E_{\bar{\mathbf{s}},\bar{\mathbf{a}} \sim \pi}\left[Q^{\pi}(\bar{\mathbf{s}},\bar{\mathbf{a}})\right]\right|, \label{eq:normalized_values_bias}
\end{align}
where
\begin{itemize} [leftmargin=8mm]
\setlength\itemsep{0.1em}
  \item $Q^{\pi}(\mathbf{s},\mathbf{a})$ be the action-value function for policy $\pi$ using the standard infinite-horizon discounted Monte Carlo return definition. 
  \item $\bar Q_{\theta }(\mathbf{s},\mathbf{a})$ the estimated Q-value, defined as the mean of $Q_{\theta _{i}}(\mathbf{s}, \mathbf{a}), i=1, \ldots, N$.
\end{itemize}

For RAC, the normalized values bias is defined as:
\begin{align}
  & \left(\bar Q_{\theta }(\mathbf{s}, \mathbf{a},\beta^*) - Q^{\pi^*}(\mathbf{s},\mathbf{a})\right) /\left|E_{\bar{s}, \bar{a} \sim \pi^*}\left[Q^{\pi^*}(\bar{\mathbf{s}}, \bar{\mathbf{a}})\right]\right|, \label{eq:rac_normalized_values}
\end{align}
where
\begin{itemize} [leftmargin=8mm]
\setlength\itemsep{0.1em}
  \item $\pi^*$ is the best-performing policy in the evaluation among $H$ policies~\ref{appendix:evaluationmethod}.
  \item $Q^{\pi^*}(\mathbf{s}, \mathbf{a})$ be the action-value function for policy $\pi^*$ using the standard infinite-horizon discounted Monte Carlo return definition.
  \item $\bar Q_{\theta }(\mathbf{s},\mathbf{a}, \beta^*)$ the estimated Q-value using $Q_{\theta }$ of $\beta^*$ which corresponds to the policy $\pi^*$, defined as the mean of $Q_{\theta _{i}}(\mathbf{s},\mathbf{a},\beta^*), i=1, \ldots, N$.
\end{itemize}

To get various target state-action pairs, we first execute the policy in the environment to obtain 100 state-action pairs and then sample the target state-action pair from them without repetition. Starting from the target state-action pair, running the Monte Carlo processes until the max step limit is reached. Table~\ref{table:perparameters_MC} lists common parameters of the normalized values bias estimation.
\begin{table}[H]
\renewcommand{\arraystretch}{1.1}
\centering
\caption{Parameters of the normalized values bias estimation}
\label{table:perparameters_MC}
\vspace{1mm}
\begin{tabular}{l l| l }
    \toprule
    \multicolumn{2}{l|}{Parameter} &  Value\\
    \midrule
        & number of Monte Carlo process  & 20\\
        & number of target state-action pairs & 20\\
        & Max step limit & 1500\\
    \end{tabular}
\end{table}

\section{Hyperparameters and implementation details} \label{appendix:hyperparameters}

We implement all RAC algorithms with Pytorch~\cite{paszke2019pytorch} and use Ray[tune]~\cite{liaw2018tune} to build and run distributed applications.
For all the algorithms and variants, we first obtain $5000$ data points by randomly sampling actions from the action space without making any parameter updates. In order to stabilize early learning of critics, a linear learning rate warm up strategy is applied to critics in the start stage of training for RAC and its variants:
\begin{align}
  & l = l_{init}(1-p)+p \cdot l_{target}, p = \operatorname{clip}(\frac{t-t_{start}}{t_{target}-t_{start}} , 0, 1), \label{eq:lr_warm_up}
\end{align}
where $l$ is current learning rate, $l_{init}$ is the initial value of learning rate, $l_{target}$ is the target value of learning rate, $t_{start}$ is the time steps to start adjusting the learning rate, $t_{target}$ is the time steps to arrive at $l_{target}$.

For all RAC algorithms and variants, We parameterize both the actor and critics with feed-forward neural networks with $256$ and $256$ hidden nodes respectively, with rectified linear units (ReLU)~\cite{nair2010rectified} between each layer. $\beta$ is log-scaled before input into actors and critics. In order to prevent $\beta$ sample from being zero, a small value $\varepsilon=10^{-7}$ is added to the left side of $U_1$ and $U_2$ to be $U_1[\varepsilon, a]$ and $U_2[\varepsilon, b]$. Weights of all networks are initialized with Kaiming Uniform Initialization~\cite{he2015delving}, and biases are zero-initialized. For all environments, we normalize actions to a range of $[-1, 1]$.

\subsection{RAC-SAC algorithm} \label{appendix:rac-sac}
Here, the policy is modeled as a Gaussian with mean and covariance given by neural networks to handle continuous action spaces. The way RAC optimize the policy makes use of the reparameterization trick~\cite{kingma2013auto, haarnoja2018soft}, in which a sample from is drawn by computing a deterministic function of the state, policy parameters, and independent noise:
\begin{align}
  & \tilde{\mathbf{a}}_{\phi}(\mathbf{s}, \beta, \xi)=\tanh \left(\mu_{\phi}(\mathbf{s}, \beta)+\sigma_{\phi}(\mathbf{s}, \beta) \odot \xi\right), \quad \xi \sim \mathcal{N}(0, I). \label{eq:reparameterization_trick}
\end{align}
The actor network outputs the Gaussian’s means and log-scaled covariance, and the log-scaled covariance is clipped in a range of $[-10, 2]$ to avoid extreme values. The actions are bounded to a finite interval by applying an invertible squashing function ($tanh$) to the Gaussian samples, and the log-likelihood of actions is calculated by the Squashed Gaussian Trick~\cite{haarnoja2018soft}.

The temperature is parameterized by a one layer feedforward neural network $T_{\psi}$ of $64$ with rectified linear units (ReLU). To prevent temperature be negative, we parameterize temperature as:
\begin{align}
  & \alpha_{\psi}(\beta) = e^{T_{\psi}(\operatorname{log}(\beta))+\xi}, \label{eq:parameterized_temp}
\end{align}
where $\xi$ is constant controlling the initial temperature, $\operatorname{log}(\beta)$ is log-scaled $\beta$, $T_{\psi}(\operatorname{log}(\beta))$ is the output of the neural network.

\subsection{RAC-TD3 algorithm} \label{appendix:rac-td3}
We implement RAC-TD3 referring to \url{https://github.com/sfujim/TD3}. A final $tanh$ unit following the output of the actor. Different from TD3, we did not use a target network for actor and delayed policy updates. For each update of critics, a small amount of random noise is added to the policy and averaging over mini-batches:
\begin{align}
  & y=r+\gamma\left[\bar{Q}_{\bar \theta}(\mathbf{s'}, \mathbf{a'},\beta)-\beta \hat{s}({Q}_{\bar \theta}(\mathbf{s'}, \mathbf{a'},\beta))\right], \label{eq:rac_td3_critic_target}
  \\
  & \mathbf{a'} = \operatorname{clip}(\pi_{\phi}\left(\cdot \mid \mathbf{s'},\beta\right)+\epsilon, -1, 1),
  \label{eq:rac_td3_policy}
    \\
  & \epsilon \sim \operatorname{clip}(\mathcal{N}(0, \sigma),-c, c). \label{eq:rac_td3_epsilon}
\end{align}
The extended policy $\pi_{\phi}$ is updated by minimizing the following object:
\begin{align}
  & \mathcal{L}^{\tt RAC-TD3}_{\tt actor} (\phi) = \mathbb{E}_{\mathbf{s} \sim \mathcal{B},\beta \sim U_{1}}\left[-\bar Q_{\theta}\left(\mathbf{a}, \mathbf{s}, \beta\right)\right], 
  \mathbf{a} = \pi_{\phi}\left(\cdot \mid \mathbf{s},\beta\right). \label{eq:rac_td3_actorot}
\end{align}
The pseudocode for RAC-TD3 is shown in Algorithm~\ref{alg:rac_td3}.

\begin{algorithm}[tb]
\caption{RAC: TD3 version} \label{alg:rac_td3}
\begin{algorithmic}[1]
\State Initialize actor network $\phi$, $N$ critic networks $\theta_i, i=1,\ldots,N$, empty replay buffer $\mathcal{B}$, target network $\bar{\theta_{i}}\longleftarrow \theta_{i}$, for $i=1,2,\ldots,N$, uniform distribution $\mathcal{U}_{1}$ and $U_{2}$
\For{each iteration}
\State execute an action $\mathbf{a} = \pi_{\phi}\left(\cdot \mid \mathbf{s},\beta\right)+\epsilon,  \epsilon \sim \mathcal{N}(0, \sigma), \beta \sim U_{2}$. Observe reward $r$, new state $\mathbf{s'}$
\State Store transition tuple $\mathcal{B} \leftarrow \mathcal{B} \cup\left\{\left(\mathbf{s}, \mathbf{a}, r, \mathbf{s'}\right)\right\}$
\For{$G$ updates}
\State Sample random minibatch  $\{\tau_j\}_{j=1}^B\sim\mathcal{B}$, $\{\beta_m\}_{m=1}^B\sim U_{1}$
\State Compute the Q target~\eqref{eq:rac_td3_critic_target}
\For{$i=1,...,N$}
\State Update $\theta_i$ by minimize $\mathcal{L}^{\tt RAC}_{\tt critic}$~\eqref{eq:rac_critic_tot_beta}
\renewcommand\baselinestretch{1.2}\selectfont
\State Update target networks with $\bar{\theta_{i}} \leftarrow \rho \bar{\theta_{i}}+(1-\rho) \theta_{i}$
\EndFor 
\renewcommand\baselinestretch{1}\selectfont
\EndFor 
\State Update actor parameters $\phi$ by minimize $\mathcal{L}^{\tt RAC-TD3}_{\tt actor} $~\eqref{eq:rac_td3_actorot}
\EndFor
\end{algorithmic}
\end{algorithm}

\subsection{Vanilla RAC algorithm} \label{appendix:vanilla_rac}
\begin{figure*}[ht]
  \centering
  \includegraphics[width=\textwidth]{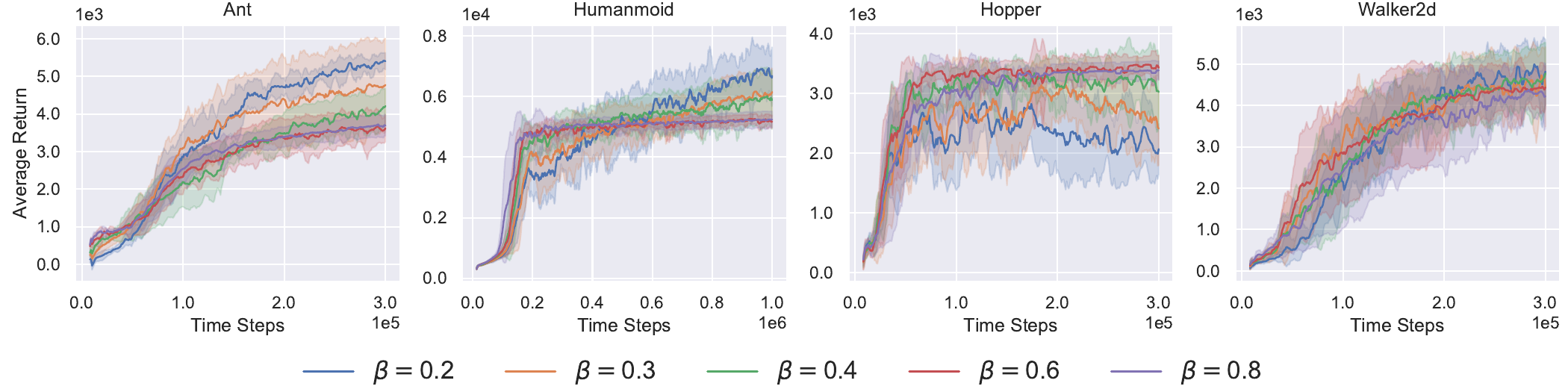}
  \caption{Performance of Vanilla RAC with different $\beta$.}
  \label{VRAC_ablation}
\end{figure*}

\begin{figure*}[ht]
  \centering
  \includegraphics[width=\textwidth]{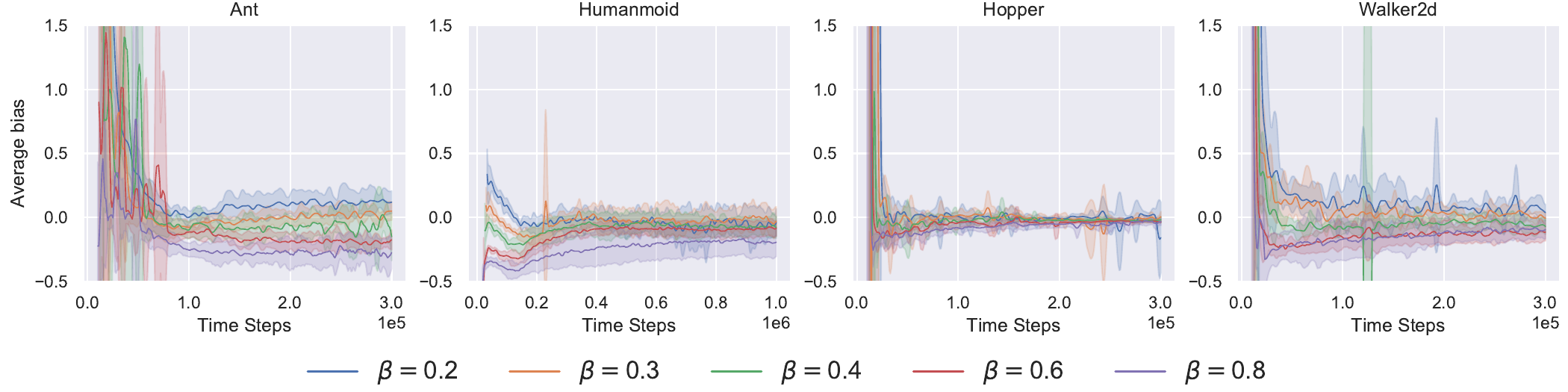}
  \caption{Mean of normalized Q bias for Vanilla RAC with different $\beta$.}
  \label{VRAC_bias}
\end{figure*}

\begin{figure*}[ht]
  \centering
  \includegraphics[width=\textwidth]{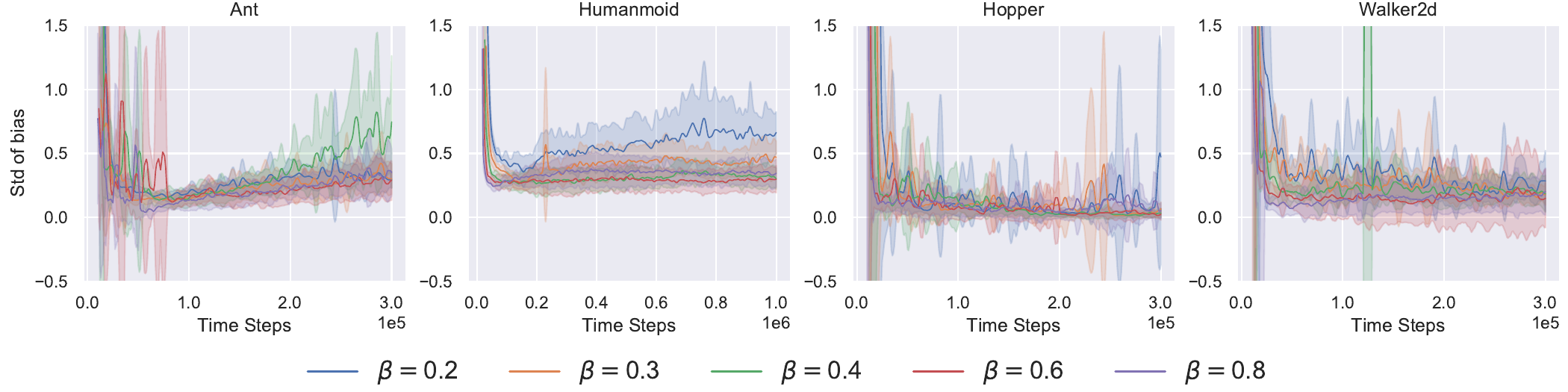}
  \caption{Std of normalized Q bias for Vanilla RAC with different $\beta$.}
  \label{VRAC_std}
\end{figure*}

For vanilla RAC, UVFA is not needed as $\beta$ is a constant. The actor is updated by minimizing the following object:
\begin{align}
  & \mathcal{L}^{\tt {vanilla RAC}}_{\tt actor} (\phi) = \mathbb{E}_{\mathbf{s} \sim \mathcal{B}, \mathbf{a} \sim \pi_{\phi}}\left[\alpha \log \left(\pi_{\phi}\left(\mathbf{a} \mid \mathbf{s}\right)\right)-\bar{Q}_{\theta}\left(\mathbf{a}, \mathbf{s}\right)\right]. \label{eq:Vanilla_actorot}
\end{align}
The pseudocode for Vanilla RAC is shown in Algorithm~\ref{alg:vanilla_rac}. Figture \ref{VRAC_ablation}, \ref{VRAC_bias}, and \ref{VRAC_std} shows performance and Q bias of Vanilla RAC with different $\beta$.

\begin{algorithm}[tb]
\caption{Vanilla RAC} \label{alg:vanilla_rac}
\begin{algorithmic}[1]
\State Initialize actor network $\phi$, $N$ critic networks $\theta_i, i=1,\ldots,N$, empty replay buffer $\mathcal{B}$, target network $\bar{\theta_{i}}\longleftarrow \theta_{i}$, for $i=1,2,\ldots,N$
\For{each iteration}
\State execute an action $\mathbf{a} \sim \pi_{\phi}\left(\cdot \mid s \right)$. Observe reward $r$, new state $\mathbf{s'}$
\State Store transition tuple $\mathcal{B} \leftarrow \mathcal{B} \cup\left\{\left(\mathbf{s}, \mathbf{a}, r, \mathbf{s'}\right)\right\}$
\For{$G$ updates}
\State Sample random minibatch  $\{\tau_j\}_{j=1}^B\sim\mathcal{B}$
\State Compute the Q target~\eqref{eq:rac_critic_target}
\For{$i=1,\ldots,N$}
\State Update $\theta_i$ by minimize $\mathcal{L}_{\tt critic}$~\eqref{eq:sac_critic_tot}
\renewcommand\baselinestretch{1.2}\selectfont
\State Update target networks with $\bar{\theta_{i}} \leftarrow \rho \bar{\theta_{i}}+(1-\rho) \theta_{i}$
\EndFor 
\renewcommand\baselinestretch{1.1}\selectfont
\EndFor 
\State Update actor parameters $\phi$ by minimize $\mathcal{L}^{\tt vanilla RAC}_{\tt actor} $~\eqref{eq:Vanilla_actorot}
\State Update temperature $\alpha$ by minimize $ \mathcal{L}_{\tt temp}$~\eqref{eq:sac_temp_loss}
\EndFor
\end{algorithmic}
\end{algorithm}

\subsection{RAC with in-target minimization} \label{appendix:RAC_in_target_minimization}

We implement RAC with in-target minimization referring to authors’s code \url{https://github.com/watchernyu/REDQ}. The critics and actor are extended as $Q_{\theta_{i}}(\mathbf{s},\mathbf{a},k)$ and $\pi_{\phi}\left(\cdot \mid \mathbf{s'}, k \right)$, $U_{1}$ is a uniform traning distribution $\mathcal{U}[1, a]$, $a>1$, $k \sim U_{1}$ that determine the size of the random subset $\mathcal{M}$. When $k$ is not an integer, the size of $\mathcal{M}$ will be sample beweetn $\operatorname{floor}(k)$ and $\operatorname{floor}(k+1)$ according to the Bernoulli distribution $\mathcal{B}(p)$ with parameter $p = k - \operatorname{floor}(k)$, where $\operatorname{floor}$ is a round-towards-zero operator.

An independent temperature network $\alpha_{\psi}$ parameterized by $\psi$ is updated with the following object:
\begin{align}
  & \mathcal{L}^{\tt RAC}_{\tt temp} (\psi) = \mathbb{E}_{\mathbf{s} \sim \mathcal{B}, \mathbf{a} \sim \pi_{\phi}, k \sim U_{1}}\left[-\alpha_{\psi}(k) \log \pi_{\phi}\left(\mathbf{a} \mid \mathbf{s}, k \right)-\alpha_{\psi}(k) \overline{\mathcal{H}}\right]. \label{eq:rac_in_target_temp_loss}
\end{align}
In-target minimization is used to calculate the target $y$:
\begin{equation}
\begin{aligned}
  &y=r+\gamma\mathbb{E}_{\mathbf{a'}\sim \pi_\phi} \left[\min _{i \in \mathcal{M}} Q_{\bar{\theta_i}}\left(\mathbf{s'}, \mathbf{a'}, k\right)-\alpha_{\psi}(k) \log \pi_{\phi}\left(\mathbf{a'} \mid \mathbf{s'}, k\right)\right], \label{eq:rac_in_target_critic_target}
\end{aligned}
\end{equation}
Then each $Q_{\theta_{i}}(\mathbf{s},\mathbf{a}, k)$ is updated with the same target:
\begin{align}
  & \mathcal{L}^{\tt RAC}_{\tt critic} (\theta_i) = \mathbb{E}_{\tau_t \sim \mathcal{B}, k \sim U_{1}} [\left(Q_{\theta_{i}}(\mathbf{s},\mathbf{a},k) - y \right)^2]. \label{eq:rac_in_target_critic_tot}
\end{align}
The extended policy $\pi_{\phi}$ is updated by minimizing the following object:
\begin{align}
  & \mathcal{L}^{\tt RAC}_{\tt actor} (\phi) = \mathbb{E}_{\mathbf{s} \sim \mathcal{B}, k \sim U_{1}}\left[\mathbb{E}_{\mathbf{a} \sim \pi_{\phi}}\left[\alpha_{\psi}(k) \log \left(\pi_{\phi}\left(\mathbf{a} \mid \mathbf{s}, k\right)\right)-\bar Q_{\theta}\left(\mathbf{a}, \mathbf{s}, k\right)\right]\right]. \label{eq:rac_in_target_actor}
\end{align}
When interacting with the environment, obtaining exploration behavior by sample $k$ from exploration distribution $U_2=\mathcal{U}[1, b]$, $a>b>1$.

The pseudocode for RAC with in-target minimization is shown in Algorithm~\ref{alg:RAC_in_target_minimization}.

\begin{algorithm}[t]
\caption{RAC with in-target minimization} \label{alg:RAC_in_target_minimization}
\begin{algorithmic}[1]
\State Initialize actor network $\phi$, $N$ critic networks $\theta_i, i=1,\ldots,N$, temperature network $\psi$, empty replay buffer $\mathcal{B}$, target network $\bar{\theta_{i}}\longleftarrow \theta_{i}$, for $i=1,2,\ldots,N$, uniform distribution $U_{1}$ and $U_{2}$
\For{each iteration}
\State execute an action $\mathbf{a} \sim \pi_{\phi}\left(\cdot \mid \mathbf{s}, k \right), k \sim U_{2}$. Observe reward $r$, new state $\mathbf{s'}$
\State Store transition tuple $\mathcal{B} \leftarrow \mathcal{B} \cup\left\{\left(\mathbf{s}, \mathbf{a}, r, \mathbf{s'}\right)\right\}$
\For{$G$ updates}
\State Sample random minibatch  $\{\tau_j\}_{j=1}^B\sim\mathcal{B}$, $\{k_j\}_{j=1} \sim U_{1}$
\State Sample a set $\mathcal{M}$ of $k$ distinct indices from $\{1,2, \ldots, N\}$
\State Compute the Q target~\eqref{eq:rac_in_target_critic_target}
\For{$i=1,\ldots,N$}
\State Update $\theta_i$ by minimize $\mathcal{L}^{\tt RAC}_{\tt critic}$~\eqref{eq:rac_in_target_critic_tot}
\renewcommand\baselinestretch{1.2}\selectfont
\State Update target networks with $\bar{\theta_{i}} \leftarrow \rho \bar{\theta_{i}}+(1-\rho) \theta_{i}$
\EndFor 
\renewcommand\baselinestretch{1.1}\selectfont
\EndFor 
\State Update actor parameters $\phi$ by minimize $\mathcal{L}^{\tt RAC-SAC}_{\tt actor} $~\eqref{eq:rac_in_target_actor}
\State Update temperature parameters $\psi$ by minimize $\mathcal{L}^{\tt RAC}_{\tt temp}$~\eqref{eq:rac_in_target_temp_loss}
\EndFor
\end{algorithmic}
\end{algorithm}

\subsection{Selection of hyperparameters} \label{appendix:hyperparameters_select}
In order to select the hyperparameters used for RAC for all mojoco environments, which are shown in table~\ref{table:hyperparameters_setting}, we ran a grid search with the ranges shown on table~\ref{table:hyperparameters_sweeps}, and the combination with the highest cumulative rewards amount environments are selected.
\begin{table}[H]
\renewcommand{\arraystretch}{1.1}
\centering
\caption{Range of hyperparameters sweeps}
\label{table:hyperparameters_sweeps}
\vspace{1mm}
\begin{tabular}{l l| l }
    \toprule
    \multicolumn{2}{l|}{Hyperparameters} &  Value\\
    \midrule
    \multicolumn{2}{l|}{\it{Shared}}& \\
        & Update-To-Data (UTD) ratio ($G$) & \{1, 5, 10, 20\}\\
        & ensemble size ($N$) & \{2, 5, 10\}\\
    \midrule
    \multicolumn{2}{l|}{\it{RAC-SAC}}& \\
        & initial temperature coefficient ($\xi$) & \{$-5, -2, 0$\}\\
        & right side of exploitation distribution $U_{1}$ ($a$) & \{$0.6, 0.8, 1.0, 1.2$\}\\
        & right side of exploration distribution $U_{2}$ ($b$) &\{$0.2, 0.3, 0.4, 0.6, 0.8$\}\\
        & replay buffer capacity & \{$0.5\times 10^{5}, 10^{5}, 2\times 10^{5}, 3\times 10^{5}, 10^{6}$\}\\
    \midrule
    \multicolumn{2}{l|}{\it{RAC-TD3}}& \\
        & right side of exploitation distribution $U_{1}$ ($a$) & \{$0.6, 0.8, 1.0, 1.2$\}\\
        & right side of exploration distribution $U_{2}$ ($b$) &\{$0.2, 0.3, 0.4, 0.6, 0.8$\}\\
        & replay buffer capacity & \{$0.5\times 10^{5}, 10^{5}, 2\times 10^{5}, 3\times 10^{5}, 10^{6}$\}\\
    \midrule
    \multicolumn{2}{l|}{\it{Vanilla RAC}}& \\
        & uncertainty punishment ($\beta$) & \{$0.2, 0.3, 0.4, 0.6, 0.8$\}\\
    \midrule
    \multicolumn{2}{l|}{\it{RAC with in-target minimization}}& \\
        & right side of exploitation distribution $U_{1}$ ($a$) & \{$2, 2.5, 3, 4$\}\\
        & right side of exploration distribution $U_{2}$ ($b$) &\{$1.25, 1.5, 1.75, 2$\}\\
        \bottomrule
    \end{tabular}
\end{table}

\subsection{Hyperparameter setting} \label{appendix:hyperparameters_setting}
Table~\ref{table:shared_hyperparameters} and~\ref{table:hyperparameters_setting} lists the hyperparameters for RAC and variants used in experiments. 

\begin{table}[H]
\renewcommand{\arraystretch}{1.1}
\centering
\caption{Shared hyperparameters}
\label{table:shared_hyperparameters}
\vspace{1mm}
\begin{tabular}{l l| l }
    \toprule
    \multicolumn{2}{l|}{Hyperparameters} &  Value\\
    \midrule
        & optimizer &Adam \citep{kingma2014adam}\\
        & actor learning rate & $3 \times 10^{-4}$\\
        & temperature learning rate & $3 \times 10^{-4}$\\
        & initial critic learning rate ($l_{init}$) & $3 \times 10^{-5}$\\
        & target critic learning rate ($l_{target}$) & $3 \times 10^{-4}$\\
        & time steps to start learning rate adjusting ($t_{start}$) & 5000\\
        & time steps to reach target learning rate ($t_{target}$) & $10^{4}$\\
        & number of hidden layers (for $\phi$ and $\theta_i$) & 2\\
        & number of hidden units per layer (for $\phi$ and $\theta_i$) & 256\\
        & number of hidden layers (for $T_{\psi}$) & 1\\
        & number of hidden units per layer (for $T_{\psi}$) & 64\\
        & discount ($\gamma$) &  0.99\\
        & nonlinearity & ReLU\\
        & evaluation frequence & $10^3$\\
        & minibatch size & 256\\
        & target smoothing coefficient ($\rho$) & 0.005\\
        & Update-To-Data (UTD) ratio ($G$) & 20\\
        & ensemble size ($N$) & 10\\
        & number of evaluation episodes & 10\\
        & initial random time steps & 5000\\
        & frequence of delayed policy updates & 1\\
        & log-scaled covariance clip range & $[-10, 2]$\\
        & number of discrete policies for evaluation ($H$) & 12\\
        \bottomrule
    \end{tabular}
\end{table}

\begin{table}[H]
\renewcommand{\arraystretch}{1.1}
\centering
\caption{Specific Hyperparameters}
\label{table:hyperparameters_setting}
\vspace{1mm}
\begin{tabular}{l l| l }
    \toprule
    \multicolumn{2}{l|}{Hyperparameters} &  Value\\
    \midrule
    \multicolumn{2}{l|}{\it{RAC-SAC}}& \\
        & initial temperature coefficient ($\xi$) & $-5$\\
        & exploitation distribution $U_{1}$ & $\mathcal{U}[10^{-7}, 0.8]$\\
        & exploration distribution $U_{2}$ & $\mathcal{U}[10^{-7}, 0.3]$\\    \midrule
    \multicolumn{2}{l|}{\it{RAC-TD3}}& \\
        & exploration noisy  & $\mathcal{N}(0,0.1)$\\
        & policy noisy ($\sigma$) & $0.2$\\
        & policy noisy clip ($c$) & $0.5$\\
        & traning distribution $U_{1}$ & $\mathcal{U}[10^{-7}, 0.8]$\\
        & exploration distribution $U_{2}$ & $\mathcal{U}[10^{-7}, 0.3]$\\
    \midrule
    \multicolumn{2}{l|}{\it{Vanilla RAC}}& \\
        & initial temperature & $\exp(-3)$\\
        & uncertainty punishment ($\beta$) & 0.3\\
        \midrule
    \multicolumn{2}{l|}{\it{RAC with in-target minimization}}& \\
        & initial temperature coefficient ($\xi$) & $-5$\\
        & exploitation distribution $U_{1}$ & $\mathcal{U}[1, 1.5]$\\
        & exploration distribution $U_{2}$ & $\mathcal{U}[1, 2.0]$\\
        \bottomrule
    \end{tabular}
\end{table}

\begin{table}[H]
\renewcommand{\arraystretch}{1.1}
\centering
\caption{Environment dependent hyperparameters}
\label{table:Environment_dependent}
\vspace{1mm}
\begin{tabular}{l l|cccc}
    \toprule
    \multicolumn{2}{l|}{Hyperparameters} &  Humanmoid & Walker & Ant & Hopper\\
    \midrule
    \multicolumn{2}{l|}{\it{RAC-SAC}}& \\
        & replay buffer capacity & $3\times10^5$ & $10^5$ & $2\times10^5$ & $1\times10^6$\\
    \midrule
    \multicolumn{2}{l|}{\it{RAC-TD3}}& \\
        & replay buffer capacity & $3\times10^5$ & $10^5$ & $2\times10^5$ & $1\times10^6$\\
    \midrule
    \multicolumn{2}{l|}{\it{Vanilla RAC}}& \\
        & replay buffer capacity & $10^6$ & $10^6$ & $10^6$ & $10^6$\\
        & uncertainty punishment ($\beta$) & 0.2 & 0.3 & 0.2 & 0.2\\
        \midrule
    \multicolumn{2}{l|}{\it{RAC with in-target minimization}}& \\
        & replay buffer capacity & $3\times10^5$ & $10^5$ & $2\times10^5$ & $1\times10^6$\\
        \bottomrule
    \end{tabular}
\end{table}





\section{Visualisations of learned temperatures} \label{appendix:temp_Visualisations}
\begin{figure*}[ht]
  \centering
  \includegraphics[width=\textwidth]{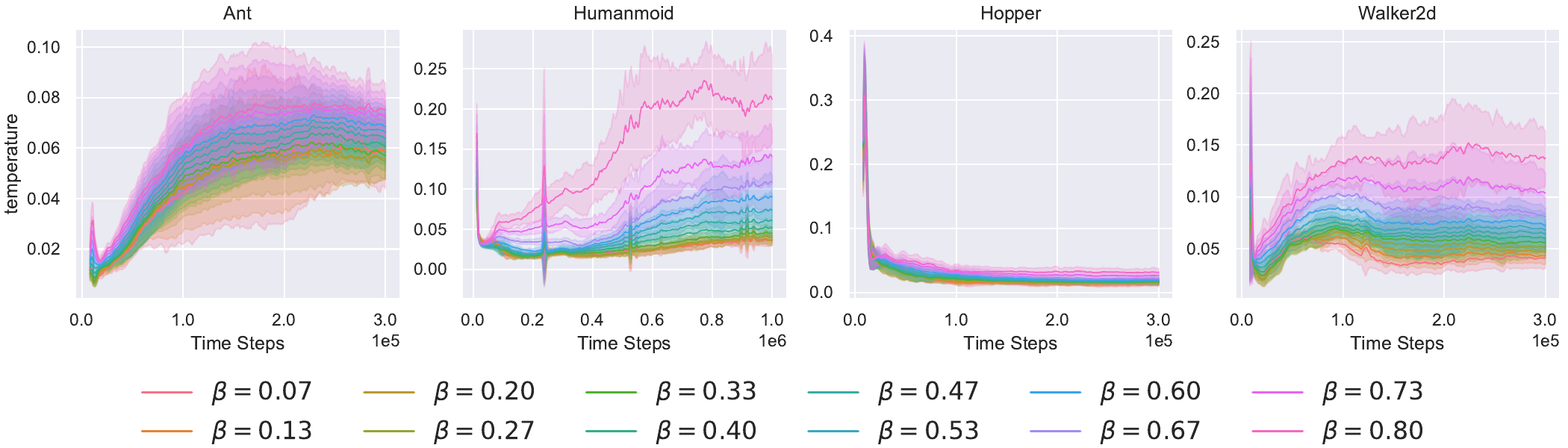}
  \caption{Visualisations of learned temperatures for RAC-SAC with different $\beta$.}
  \label{temp_Visualisations}
\end{figure*}

The figure \ref{temp_Visualisations} shows the visualization of learned temperatures with respect to different $\beta$ during training. The figure demonstrates that learned temperatures are quite different, it is difficult to take into account the temperature of different $\beta$ with a single temperature.

